\documentclass[journal]{IEEEtran}
\usepackage[figuresright]{rotating}
\usepackage{amssymb}
\usepackage{amsthm}
\usepackage{multicol}
\usepackage{subfigure}
\usepackage{graphicx}
\usepackage{epstopdf}
\usepackage{fullpage}
\usepackage{latexsym,amsmath}
\usepackage{stmaryrd}
\usepackage{algorithm,algorithmic}
\usepackage{subfigure}
\usepackage{amsfonts}
\usepackage{xcolor,multirow}
\usepackage{cite}
\usepackage{bm}
\usepackage{url}
\usepackage{diagbox}
\usepackage{array}
\usepackage{booktabs}
\usepackage{footmisc} 
\usepackage{pifont}
\usepackage{caption}

\newtheorem{theorem}{Theorem}
\newtheorem{definition}{Definition}
\newtheorem{remark}{Remark}
\newtheorem{property}{Property}
\newtheorem{corollary}{Corollary}

\ifCLASSINFOpdf
\else
\fi
\begin{document}
	
\title{Non-Negative Reduced Biquaternion Matrix Factorization with Applications in Color Face Recognition}

\author{Jifei~Miao, Junjun~Pan, and  Michael~K.~Ng, ~\IEEEmembership{Senior Member,~IEEE}
\thanks{Jifei Miao is with the School of Mathematics and Statistics, Yunnan University, Kunming, Yunnan, 650091, China (e-mail: jifmiao@163.com
	).}
\thanks{Junjun Pan and Michael K. Ng are with the Department of Mathematics, Hong Kong
	Baptist University, Kowloon Tong, Hong Kong, China (e-mail: junjpan@hkbu.edu.hk; michael-ng@hkbu.edu.hk).
	}
}

\markboth{}%
{Shell \MakeLowercase{\textit{}}: Bare Demo of IEEEtran.cls for IEEE Journals}

\maketitle

\begin{abstract}
  Quaternion numbers are commonly used to represent color image pixels. Since color image pixels are non-negative, non-negativity constraint should be incorporated into the components of quaternion numbers. However, existing quaternion-based matrix factorization models do not fully satisfy non-negativity constraints. The main purpose of this paper is to propose and develop non-negative quaternion matrix factorization models to address the relevant color image processing problems. Our main idea is to employ the product of reduced biquaternion (RB) matrices in the factorization such that non-negativity constraint can be incorporated properly. The proposed non-negative RB matrix factorization (NRBMF) model is solved by an alternating minimization algorithm, which alternately solves the RB non-negative least squares problem of latent features and encoding factors. The theoretical convergence of the alternating minimization algorithm is also established. Experimental results of color face recognition show that the recognition accuracy of the proposed NRBMF model is $4.1\%$ higher than that of the existing models on average.
\end{abstract}


\IEEEpeerreviewmaketitle

\section{Introduction}

Color image processing plays a key role in computer vision applications such as face recognition \cite{lu2018color}, medical imaging \cite{ashikuzzaman2019low}, and remote sensing \cite{kior2024rgb}. Unlike grayscale images, color images require joint representation of multiple channels (\emph{e.g.}, RGB), where correlations between channels are critical for achieving high-fidelity results \cite{cui2020color}. However, traditional methods based on real-valued matrices typically handle each channel independently or rely on simple concatenation models neglecting inter-channel dependencies and thereby resulting in suboptimal performance \cite{xu2015vector}.

To address this, quaternion algebra \cite{hamilton1866elements} has gained widespread attention as a powerful tool for representing color pixels. A quaternion comprises one real component and three imaginary components, with the RGB channels of a color pixel typically encoded in the latter three \cite{chen2019low,jia2022non}. This representation treats each color pixel as a unified entity, enabling the intrinsic relationships among color channels to be fully preserved and effectively exploited \cite{wang2021robust}. Consequently, quaternion-based color image processing has recently become a subject of extensive research, such as quaternion matrix low-rank approximation for color image inpainting and color image denoising \cite{miao2020quaternion,miao2021color,song2021low,yu2019quaternion}, quaternion sparse representation for color image recognition \cite{zou2016quaternion}, and quaternion dynamic mode decomposition for foreground-background separation in color videos \cite{han2022quaternion}.

To represent color images property, non-negativity of pixel values must be incorporated into quaterion numbers. 
This can be naturally addressed by restricting the three imaginary components (encoding RGB channels) of pure quaternion matrices (\emph{i.e.}, with zero real component) to be non-negative. This idea was demonstrated in \cite{lyu2024randomized}, which proposed a non-negative quaternion matrix low-rank approximation method that achieved promising results in color image low-rank reconstruction.

In many image processing applications, non-negative matrix factorization (NMF) \cite{lee1999learning} offers both non-negativity constraints and feature extraction capability with strong interpretability. In particular, its application in face recognition has been extensively studied \cite{guillamet2002non,wang2005non,nikan2020face,chen2022novel}. 
The basic idea is to decompose a given non-negative matrix (with non-negative real numbers) into the product of the 
two non-negative factor matrices. The factor matrices are regarded as a feature matrix and an encoding matrix respectively.   
Note that the product of two non-negative factor matrices must be a non-negative matrix which is used to approximate 
the original given non-negative matrix. 
For color image analysis, it is natural to decompose a given non-negative quaternion matrix (with non-negative quaternion numbers) into the product of two non-negative quaternion matrices. Because of non-commutativity of 
quaternion multiplication, 
the matrix entries of the product of two non-negative quaternion matrices are not necessary non-negative. 
Recent work \cite{ke2023quasi} attempted to address this issue by defining the quasi non-negative quaternion matrix constraining the three imaginary components to be non-negative while allowing unconstrained real component. 
Following this definition, they proposed a quasi non-negative quaternion matrix factorization (QNQMF) model. 
However, the QNQMF model cannot theoretically guarantee that the product of two quasi non-negative quaternion factor matrices remains quasi non-negative. In the literature, there is still no proper non-negative quaternion matrix factorization model 
that can be used for feature extraction and interpretation.

The main aim of this paper is to propose and develop non-negative 
quaternion matrix factorization models to address relevant color image 
processing problems. 
Our main idea is to employ the product of reduced biquaternion (RB) matrices \cite{segre1892real,pei2004commutative}
in the factorization such that non-negativity constraint can be incorporated properly.
The key property of RB numbers is its commutative nature of RB multiplication. 
Therefore, RB has shown considerable promise in color image processing \cite{el2017color,gai2023theory,guo2024algebraic}. 
In this work, for color image processing, we utilize the multiplication properties of RB algebra and propose an 
non-negative RB matrix factorization (NRBMF) model. 
The contributions of this paper are given as follows:
\begin{itemize}
	\item To effectively handle color images, we construct a novel NMF framework in the RB domain by introducing the concept of non-negative RB matrices and developing the NRBMF model. Specifically, non-negative RB matrices enforce non-negativity constraints on all real and imaginary components. The NRBMF model decomposes a non-negative RB matrix into two non-negative RB matrices: a basis matrix and an encoding coefficient matrix. Specially, by constraining the encoding coefficient matrix to contain only the real component and the second imaginary component, we rigorously guarantee that their product remains a valid non-negative RB matrix.
	\item We derive the steepest descent direction for the real-valued objective function with RB matrix variables, and for convenience, denote its negative as the `gradient' in subsequent derivations. Based on this compact representation, we develop an efficient RB projected gradient algorithm for the NRBMF optimization problem, accompanied by a convergence analysis.
	\item We apply the proposed NRBMF model to color face recognition.  To enhance the representational capacity of the model, we propose using a full RB matrix with a real component for color image representation, as opposed to the conventional pure RB matrix. The experimental results indicate that the method exhibits highly competitive performance.
\end{itemize}

The rest paper is organized as follows. In Section \ref{1}, we introduce non-negative RB matrices and 
the proposed NRBMF model. 
Section \ref{OAsec3} presents the projected gradient algorithm for solving the proposed model, along with convergence analysis of the algorithm. 
Section \ref{experiments} validates the effectiveness of the proposed method in color face recognition. The conclusion is ultimately provided in Section \ref{conclusion}.

\subsection{Notations}

Some basic notations used in the paper are given in Table \ref{tablenot}.
\begin{table}[ht]
	\caption{Basic notations}\label{tablenot}
	\centering
	\begin{tabular}{l|c}
		\hline\hline
		Notations & Representations \\ \hline
		$\mathbb{R}$, $\mathbb{C}$, $\mathbb{Q}$, $\mathbb{RB}$ & Real, complex, quaternion, and RB spaces\\ 
		$a$, $\mathbf{a}$, $\mathbf{A}$  &  Scalar, vector, and matrix \\
		$\dot{a}$,  $\dot{\mathbf{a}}$, $\dot{\mathbf{A}}$& Quaternion scalar, vector, and matrix\\
		$\ddot{a}$,  $\ddot{\mathbf{a}}$, $\ddot{\mathbf{A}}$& RB scalar, vector, and matrix\\
		$\mathbf{i}, \mathbf{j}, \mathbf{k}$&The imaginary units of quaternions\\
		$i, j, k$& The imaginary units of RBs\\
		${\rm{Im}}_{\mathbf{i}}(\cdot)$, ${\rm{Im}}_{\mathbf{j}}(\cdot)$,  ${\rm{Im}}_{\mathbf{k}}(\cdot)$ & The three imaginary components of quaternions\\
		${\rm{Im}}_{i}(\cdot)$, ${\rm{Im}}_{j}(\cdot)$, ${\rm{Im}}_{k}(\cdot)$& The three imaginary components of RBs\\
		${\rm{Re}}(\cdot)$& The real component of quaternions or RBs\\
		$(\cdot)^{\ast}$, $(\cdot)^\top$& Conjugate and transpose\\
		$(\cdot)^{H}$, $(\cdot)^{-1}$& Conjugate transpose and inverse\\
		$\otimes$ & The element-wise product between matrices\\
		${\rm{vec}}(\cdot)$&  The vectorization operation of a matrix\\
		${\rm{cond}}(\cdot)$& The condition number of a matrix\\
		${\rm{rank}}(\cdot)$&  The rank of the matrix\\
		${\rm{abs}}(\cdot)$&  The absolute value of the input \\
		$\langle \cdot, \cdot\rangle$ & The inner product operation\\
		$|\cdot|$, $\|\cdot\|_{F}$ & Modulus and Frobenius norm	\\
		$\overset{\Delta}{=}$& Defined as\\
		$\mathbf{I}_{l}$& The identity matrix of size $l \times l$\\
		\hline\hline
	\end{tabular}
\end{table}

\section{Non-negative Bireduced Quaternion Matrix Factorization}
\label{1}

\subsection{Related Work}\label{rw}

Since color image pixels are intrinsically non-negative across all three channels, preserving this property while maintaining inter-channel correlations is essential for feature extraction tasks like color face recognition. 
To achieve this, Ke \emph{et al.} developed a quaternion-based NMF model in \cite{ke2023quasi}. 
A quaternion matrix  $\dot{\mathbf{Q}}=\mathbf{Q}_{0}+\mathbf{Q}_{1}\mathbf{i}+\mathbf{Q}_{2}\mathbf{j}+\mathbf{Q}_{3}\mathbf{k}$ is called the QNQM if $\mathbf{Q}_{1}$, $\mathbf{Q}_{2}$, and $\mathbf{Q}_{3}$ are real non-negative matrices, that is
\begin{equation*}\small
	\mathbf{Q}_{1}\geq0,\quad 	\mathbf{Q}_{2}\geq0,\quad 	\mathbf{Q}_{3}\geq0. 	
\end{equation*}	
The set of QNQM is denoted by $\mathbb{Q}^{M\times N}_{+}$. Then, following the definition of QNQM, they proposed a QNQMF model: 	For a given quaternion matrix $\dot{\mathbf{X}}=\mathbf{X}_{0}+\mathbf{X}_{1}\mathbf{i}+\mathbf{X}_{2}\mathbf{j}+\mathbf{X}_{3}\mathbf{k}\in\mathbb{Q}^{M\times N}_{+}$, QNQMF finds the matrices  $\dot{\mathbf{W}}=\mathbf{W}_{0}+\mathbf{W}_{1}\mathbf{i}+\mathbf{W}_{2}\mathbf{j}+\mathbf{W}_{3}\mathbf{k}\in\mathbb{Q}^{M\times l}_{+}$ and $\dot{\mathbf{H}}=\mathbf{H}_{0}+\mathbf{H}_{1}\mathbf{i}+\mathbf{H}_{2}\mathbf{j}+\mathbf{H}_{3}\mathbf{k}\in\mathbb{Q}^{l\times N}_{+}$ such that
\begin{equation*}\small\label{QNQMF}
	\dot{\mathbf{X}}=\dot{\mathbf{W}}\dot{\mathbf{H}},
\end{equation*}
that is
\begin{equation}\small\label{QNQMF2}
	\begin{split}
		\dot{\mathbf{X}}
		=&(\mathbf{W}_{0}\mathbf{H}_{0}-\mathbf{W}_{1}\mathbf{H}_{1}-\mathbf{W}_{2}\mathbf{H}_{2}-\mathbf{W}_{3}\mathbf{H}_{3})\\
		&+(\mathbf{W}_{0}\mathbf{H}_{1}+\mathbf{W}_{1}\mathbf{H}_{0}+\mathbf{W}_{2}\mathbf{H}_{3}-\mathbf{W}_{3}\mathbf{H}_{2})\mathbf{i}\\
		&+(\mathbf{W}_{0}\mathbf{H}_{2}-\mathbf{W}_{1}\mathbf{H}_{3}+\mathbf{W}_{2}\mathbf{H}_{0}+\mathbf{W}_{3}\mathbf{H}_{1})\mathbf{j}\\
		&+(\mathbf{W}_{0}\mathbf{H}_{3}+\mathbf{W}_{1}\mathbf{H}_{2}-\mathbf{W}_{2}\mathbf{H}_{1}+\mathbf{W}_{3}\mathbf{H}_{0})\mathbf{k}.
	\end{split}
\end{equation}
However, from (\ref{QNQMF2}), one can find that even $\dot{\mathbf{W}}\in\mathbb{Q}^{M\times l}_{+}$ and $\dot{\mathbf{H}}\in\mathbb{Q}^{l\times N}_{+}$, (\emph{i.e.}, $\mathbf{W}_{1}\geq0,	\mathbf{W}_{2}\geq0, \mathbf{W}_{3}\geq0, \mathbf{H}_{1}\geq0, \mathbf{H}_{2}\geq0, \mathbf{H}_{3}\geq0$), it is hard to guarantee that
$\dot{\mathbf{W}}\dot{\mathbf{H}}\in\mathbb{Q}^{M\times N}_{+}$ (\emph{i.e.}, each imaginary component on the right 
hand side of equation (\ref{QNQMF2})  is difficult to guarantee as non-negative). Besides, the definition of QNQM does not restrict the non-negativity of the real component of the quaternion matrix, which will make the real component lack interpretability. In specific applications, such as color face recognition, the real component of the basis matrix $\dot{\mathbf{W}}$ and encodings $\dot{\mathbf{H}}$ containing negative entries cannot be reasonably explained or utilized.

In the literature, other studies have also attempted to define non-negativity and NMF in the quaternion domain. For example, the authors in \cite{flamant2020quaternion} defined a set of non-negative quaternion matrices as: $\mathbb{Q}_{S} \overset{\Delta}{=}\{\dot{q}\in\mathbb{Q}|{\rm{Re}}(\dot{q})\geq0,{\rm{Im}}_{\mathbf{i}}(\dot{q})^{2}+{\rm{Im}}_{\mathbf{j}}(\dot{q})^{2}+{\rm{Im}}_{\mathbf{k}}(\dot{q})^{2}\leq{\rm{Re}}(\dot{q})^{2}\}$. However, this formulation was specifically developed for polarized signals, where quaternion matrices represent Stokes vectors. For color images, the channel characteristics typically violate the constraints of $\mathbb{Q}_{S}$, making this definition unsuitable.

In summary, for color image processing tasks, no existing quaternion-based NMF model both defines non-negativity in a way that reflects the characteristics of color image pixels and preserves it during the factorization process. 
This serves as the primary motivation for
proposing an NMF model based on the RB algebra in
this paper. 

\subsection{Reduced Biquaternion}
\label{a_qsec1}

A reduced biquaternion (RB) number $\ddot{q} \in\mathbb{RB}$ is defined as \cite{schutte1990hypercomplex,dimitrov1992multiplication,pei2004commutative}
\begin{equation}\small
	\label{aequ1}
	\ddot{q}=q_{0}+q_{1}i+q_{2}j+q_{3}k,
\end{equation}
where $q_{l}\in\mathbb{R}\: (l=0,1,2,3)$, and $i, j, k$ are
imaginary units satisfying
\begin{equation}\small\label{rbrules}
	\left\{
	\begin{array}{lc}
		i^{2}=k^{2}=-j^{2}=-1,\\
		ij=ji=k, jk=kj=i, ki=ik=-j.
	\end{array}
	\right.
\end{equation}
The rules outlined in (\ref{rbrules}) guarantee that the multiplication of two RB numbers is commutative, which stands in sharp contrast to the non-commutative nature of quaternion multiplication. This distinction arises because the imaginary units of quaternions satisfy the following algebraic relations \cite{hamilton1866elements}:
\begin{equation}\small
	\left\{
	\begin{array}{lc}
		\mathbf{i}^{2}=\mathbf{j}^{2}=\mathbf{k}^{2}=-1,\\
		\mathbf{i}\mathbf{j}=-\mathbf{j}\mathbf{i}=\mathbf{k}, \mathbf{j}\mathbf{k}=-\mathbf{k}\mathbf{j}=\mathbf{i}, \mathbf{k}\mathbf{i}=-\mathbf{i}\mathbf{k}=\mathbf{j}.
	\end{array}
	\right.
\end{equation}
It is important to note that the commutativity of the multiplication between RB imaginary units, along with the condition $j^{2}=1$, distinguishes RB from quaternion algebra and enables the formulation of NMF within the RB domain in this work.

The conjugate and modulus of an RB number $\ddot{q} \in\mathbb{RB}$ are defined as \cite{pei2008eigenvalues, el2022linear}
\begin{equation*}\small
	\ddot{q}^{*}=q_{0}-q_{1}i+q_{2}j-q_{3}k,	\quad 
	|\ddot{q}|=\sqrt{q_{0}^{2}+q_{1}^{2}+q_{2}^{2}+q_{3}^{2}}.
\end{equation*} 
Note that the RB system is not a complete division system \cite{pei2008eigenvalues}. However, this has almost no influence on signal and image processing applications \cite{pei2004commutative}. 

An RB matrix $\ddot{\mathbf{Q}}=(\ddot{q}_{mn})\in\mathbb{RB}^{M\times N}$ is given
by
\begin{equation*}\small
	\ddot{\mathbf{Q}}=\mathbf{Q}_{0}+\mathbf{Q}_{1}i+\mathbf{Q}_{2}j+\mathbf{Q}_{3}k,
\end{equation*}
where $\mathbf{Q}_{l}\in\mathbb{R}^{M\times N}\: (l=0,1,2,3)$. $\ddot{\mathbf{Q}}^\top=(\ddot{q}_{nm})\in\mathbb{RB}^{N\times M}$, $\ddot{\mathbf{Q}}^{*}=(\ddot{q}_{mn}^{*})\in\mathbb{RB}^{M\times N}$, and $\ddot{\mathbf{Q}}^{H}=(\ddot{q}_{nm}^{*})\in\mathbb{RB}^{N\times M}$ respectively represent the transpose, conjugate, and conjugate transpose of $\ddot{\mathbf{Q}}=(\ddot{q}_{mn})\in\mathbb{RB}^{M\times N}$. If $\mathbf{Q}_{0}=0$, the matrix $\ddot{\mathbf{Q}}$  is referred to as a pure RB matrix. The inner product and the Frobenius norm of RB matrices are defined as follows:
\begin{equation*}\small
\begin{split}
	&\langle \ddot{\mathbf{Q}},\ \ddot{\mathbf{P}}\rangle=\sum_{m=1}^{M}\sum_{n=1}^{N}\ddot{q}_{mn}^{*}\ddot{p}_{mn},\\
	&\|\ddot{\mathbf{Q}}\|_{F}=\sqrt{{\rm{Re}}(\langle\ddot{\mathbf{Q}},\ \ddot{\mathbf{Q}}\rangle)}=\sqrt{\sum_{m=1}^{M}\sum_{n=1}^{N}|\ddot{q}_{mn}|^{2}}.
\end{split}
\end{equation*}

\begin{definition} (Non-Negative RB Matrix) An RB matrix $\ddot{\mathbf{Q}}=\mathbf{Q}_{0}+\mathbf{Q}_{1}i+\mathbf{Q}_{2}j+\mathbf{Q}_{3}k$ is called a non-negative RB matrix if $\mathbf{Q}_{l}\: (l=0,1,2,3)$ are real non-negative matrices, that is
	\begin{equation*}\small
		\mathbf{Q}_{0}\geq0,\quad 	\mathbf{Q}_{1}\geq0,\quad 	\mathbf{Q}_{2}\geq0,\quad 	\mathbf{Q}_{3}\geq0. 	
	\end{equation*}	
	The set of non-negative RB matrices is denoted by $\mathbb{RB}^{M\times N}_{+}$. Specially, let the set composed of non-negative RB matrices with the first and third imaginary components being zero be denoted as $\mathbb{RB}^{M\times N}_{+j}$, i.e., if $\ddot{\mathbf{Q}}\in\mathbb{RB}^{M\times N}_{+j}$, then $\ddot{\mathbf{Q}}$ can be written as 
	\begin{equation*}\small
		\ddot{\mathbf{Q}}=\mathbf{Q}_{0}+\mathbf{Q}_{2}j,	
	\end{equation*}	
	where $\mathbf{Q}_{0}\geq0$ and	$\mathbf{Q}_{2}\geq0$.
\end{definition}

\subsection{The Proposed Model}

Based on the non-negative RB matrix definition, we propose the following 
non-negative RB matrix factorization model, which is used to address the limitations of existing 
quaternion-based NMF models in handling color images.

\begin{definition} (NRBMF)
	For a given RB matrix $\ddot{\mathbf{X}}=\mathbf{X}_{0}+\mathbf{X}_{1}i+\mathbf{X}_{2}j+\mathbf{X}_{3}k\in\mathbb{RB}^{M\times N}_{+}$, NRBMF is to find two RB matrices $\ddot{\mathbf{W}}=\mathbf{W}_{0}+\mathbf{W}_{1}i+\mathbf{W}_{2}j+\mathbf{W}_{3}k\in\mathbb{RB}^{M\times l}_{+}$ and $\ddot{\mathbf{H}}=\mathbf{H}_{0}+\mathbf{H}_{2}j\in\mathbb{RB}^{l\times N}_{+j}$ such that
	\begin{equation}\small\label{NRBMF}
		\ddot{\mathbf{X}}=\ddot{\mathbf{W}}\ddot{\mathbf{H}},
	\end{equation}
	that is
	\begin{equation}\small\label{NRBMF2}
		\begin{split}
			\mathbf{X}_{0}+\mathbf{X}_{1}i+\mathbf{X}_{2}j+\mathbf{X}_{3}k	
			=&(\mathbf{W}_{0}\mathbf{H}_{0}+\mathbf{W}_{2}\mathbf{H}_{2})\\
			&+(\mathbf{W}_{1}\mathbf{H}_{0}+\mathbf{W}_{3}\mathbf{H}_{2})i\\
			&+(\mathbf{W}_{0}\mathbf{H}_{2}+\mathbf{W}_{2}\mathbf{H}_{0})j\\
			&+(\mathbf{W}_{1}\mathbf{H}_{2}+\mathbf{W}_{3}\mathbf{H}_{0})k,
		\end{split}
	\end{equation}
	where $l$ is a pre-specified positive integer with $l\leq\min(M,N)$.
\end{definition}

In the model, $\ddot{\mathbf{W}}$ can be considered as a basis matrix,
		where $\mathbf{W}_{0}$, $\mathbf{W}_{1}$, $\mathbf{W}_{2}$, and $\mathbf{W}_{3}$ can be naturally interpreted as the four channels of the basis. $\mathbf{H}_{0}$ and $\mathbf{H}_{2}$ can be regarded as a set of combined encoding coefficient matrices.		 
		We remark the structure of $\ddot{\mathbf{H}}$ cannot be a full non-negative RB matrix containing all four components, nor can it take any form other than $\ddot{\mathbf{H}}=\mathbf{H}_{0}+\mathbf{H}_{2}j\in\mathbb{RB}^{l\times N}_{+j}$, as this may result in $\ddot{\mathbf{W}}\ddot{\mathbf{H}}\notin\mathbb{RB}^{M\times N}_{+}$, which can be directly confirmed by the multiplication rules of RB.
		
In general, the NRBMF model is not unique. Similar to NMF, it exhibits both scale and permutation ambiguities, as formally stated in Theorem \ref{them3} below.

\begin{theorem}\label{them3}
	Let $\ddot{\mathbf{X}}\in\mathbb{RB}^{M\times N}_{+}$ be a non-negative RB matrix, and let $l\leq\min(M,N)$. If $\ddot{\mathbf{X}}$ admits an NRBMF:
		\begin{equation*}\small
			\ddot{\mathbf{X}}=\ddot{\mathbf{W}}\ddot{\mathbf{H}}, \quad \ddot{\mathbf{W}}\in\mathbb{RB}^{M\times l}_{+}, \ \ddot{\mathbf{H}} \in\mathbb{RB}^{l\times N}_{+j},
		\end{equation*}
		then the factorization is non-unique up to:
		\begin{enumerate}
			\item \textbf{Scale ambiguity}: For any positive diagonal matrix $\mathbf{D} = \operatorname{diag}(d_1, \ldots, d_l)$ with $d_i > 0$, the pair $(\ddot{\mathbf{W}}\mathbf{D}, \mathbf{D}^{-1}\ddot{\mathbf{H}})$ yields an equivalent factorization:
			\begin{equation*}\small
				\ddot{\mathbf{X}} = (\ddot{\mathbf{W}}\mathbf{D})(\mathbf{D}^{-1}\ddot{\mathbf{H}}).
			\end{equation*}
			\item \textbf{Permutation ambiguity}: For any permutation matrix $\mathbf{\Pi} \in \{0,1\}^{l \times l}$ (where $\mathbf{\Pi}\mathbf{\Pi}^\top = \mathbf{I}_l$), the pair $(\ddot{\mathbf{W}}\mathbf{\Pi}, \mathbf{\Pi}^\top \ddot{\mathbf{H}})$ yields an equivalent factorization:
			\begin{equation*}\small
				\ddot{\mathbf{X}} = (\ddot{\mathbf{W}}\mathbf{\Pi})(\mathbf{\Pi}^\top \ddot{\mathbf{H}}).
			\end{equation*}
	\end{enumerate}
\end{theorem}

The conclusions in Theorem \ref{them3} follow directly and are readily verifiable.
	
\begin{figure}[htbp]
\centering
\includegraphics[width=7.8cm,height=1.8cm]{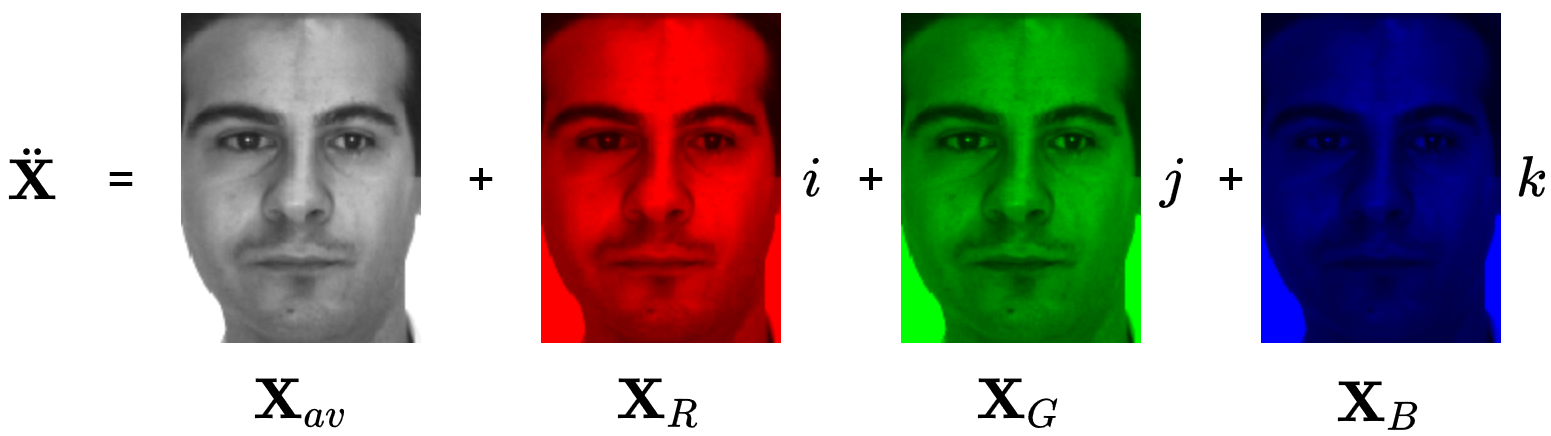} 
\caption{Color face image representation using an RB matrix.}
\label{fig_face_re}
\end{figure}

To present a color image, here we propose the following RB matrix representation: 
\begin{equation}\small\label{face_re}
	\ddot{\mathbf{X}}=\mathbf{X}_{av}+\mathbf{X}_{R}i+\mathbf{X}_{G}j+\mathbf{X}_{B}k, 
\end{equation}
where $\mathbf{X}_{av}=(\mathbf{X}_{R}+\mathbf{X}_{G}+\mathbf{X}_{B})/3$ represents the average of the three color channels instead of a zero matrix. Figure \ref{fig_face_re} visually demonstrates this representation strategy more clearly.
It should be noted that (\ref{face_re}) represents a novel approach for color image representation. The inclusion of $\mathbf{X}_{av}$, compared to (\ref{colorimage_re}), introduces no additional information, serving merely as the average of the three color channels. Thus, in practical applications, it imposes no extra requirements on the data.

\begin{remark}\label{remark1}
Note that a color face image with three channels (in RGB format, for example) and a spatial resolution of $M\times N$ can generally be represented as a pure RB matrix $\ddot{\mathbf{X}}\in\mathbb{RB}^{M\times N}$, as follows \cite{el2022linear,gai2023theory,guo2024algebraic}:
\begin{equation}\small\label{colorimage_re}
	\ddot{\mathbf{X}}=\mathbf{X}_{R}i+\mathbf{X}_{G}j+\mathbf{X}_{B}k, 
\end{equation}
where $\mathbf{X}_{R}\in\mathbb{R}^{M\times N}$, $\mathbf{X}_{G}\in\mathbb{R}^{M\times N}$, and $\mathbf{X}_{B}\in\mathbb{R}^{M\times N}$ represent the values of its red, green, and blue channels, respectively.
However, when employing a pure RB matrix encoding (i.e., ${\rm{Re}}(\ddot{\mathbf{X}})=0$), we require
the term $(\mathbf{W}_{0}\mathbf{H}_{0}+\mathbf{W}_{2}\mathbf{H}_{2})$ in (\ref{NRBMF2}) to be zero. 
Given the non-negativity constraints on $\mathbf{W}_0$, $\mathbf{W}_2$, $\mathbf{H}_0$, and $\mathbf{H}_2$, this naturally leads to the orthogonality requirement between $\mathbf{W}_0$ and $\mathbf{H}_0$, as well as between $\mathbf{W}_2$ and $\mathbf{H}_2$. These additional constraints force 
$\mathbf{W}_0$, $\mathbf{W}_2$, $\mathbf{H}_0$, and $\mathbf{H}_2$ to be more complicated in the factorization process. 
Also the orthogonality between the basis ($\mathbf{W}_0$, $\mathbf{W}_2$) and the encoding coefficients
($\mathbf{H}_0$, $\mathbf{H}_2$) lacks a physical interpretation.
According to the experimental results in Section \ref{experiments}, it is evident that the representation in (\ref{colorimage_re}) diminishes the expressive power of the factorization model.
\end{remark} 

\section{The Optimization Algorithm}\label{OAsec3}

In this section, we focus on the optimization problem of the proposed model.
\subsection{RB Optimization Problem}\label{rbopp}
To solve the NRBMF in (\ref{NRBMF}), we consider the following simple optimization problem:
\begin{equation}\small\label{opp}
	\begin{split}
		&\mathop{{\rm{min}}} f(\ddot{\mathbf{W}},\ddot{\mathbf{H}})=\frac{1}{2}\|\ddot{\mathbf{X}}-\ddot{\mathbf{W}}\ddot{\mathbf{H}}\|_{F}^{2}\\ 
		&\ \text{s.t.}\ \ddot{\mathbf{W}}\in\mathbb{RB}^{M\times l}_{+},\ \ddot{\mathbf{H}}\in\mathbb{RB}^{l\times N}_{+j}.
	\end{split}
\end{equation}
One can find that the above object function $f(\ddot{\mathbf{W}},\ddot{\mathbf{H}})$ is a real-valued function with two RB variables. Thus, the gradients of $f(\ddot{\mathbf{W}},\ddot{\mathbf{H}})$ with respect to $\ddot{\mathbf{W}}$ and $\ddot{\mathbf{H}}$ are given by (\ref{gwe}) and (\ref{ghe}) in the following.

\subsubsection{The `Gradient' of a Real Function Involving RB Matrix Variables}\label{gofrf}
Now, we will first derive the `gradient' expression of the following real-valued function with general RB matrix variables.
Suppose that $f: (\mathbb{RB}^{M\times l},\mathbb{RB}^{l\times N})\rightarrow \mathbb{R}$ be defined by
\begin{equation}\label{rbe1}
f(\ddot{\mathbf{W}},\ddot{\mathbf{H}})=\frac{1}{2}\|\ddot{\mathbf{X}}-\ddot{\mathbf{W}}\ddot{\mathbf{H}}\|_{F}^{2},
\end{equation}
where $\ddot{\mathbf{W}}=\mathbf{W}_{0}+\mathbf{W}_{1}i+\mathbf{W}_{2}j+\mathbf{W}_{3}k\in\mathbb{RB}^{M\times l}$, $\ddot{\mathbf{H}}=\mathbf{H}_{0}+\mathbf{H}_{1}i+\mathbf{H}_{2}j+\mathbf{H}_{3}k\in\mathbb{RB}^{l\times N}$, and  $\ddot{\mathbf{X}}=\mathbf{X}_{0}+\mathbf{X}_{1}i+\mathbf{X}_{2}j+\mathbf{X}_{3}k\in\mathbb{RB}^{M\times N}$. The function $f(\ddot{\mathbf{W}},\ddot{\mathbf{H}})$  can be rewritten as follows:
	\begin{equation}\label{re1}\small
 f(\ddot{\mathbf{W}},\ddot{\mathbf{H}})=\frac{1}{2}\|\ddot{\mathbf{X}}-\ddot{\mathbf{W}}\ddot{\mathbf{H}}\|_{F}^{2}
		=\frac{1}{2}\sum_{k=0}^{3}\|\mathbf{G}_{k}\|_{F}^{2},
\end{equation}
where 
\begin{equation*}\small
	\begin{split}
		\mathbf{G}_{0}=&\mathbf{X}_{0}-\mathbf{W}_{0}\mathbf{H}_{0}+\mathbf{W}_{1}\mathbf{H}_{1}-\mathbf{W}_{2}\mathbf{H}_{2}+\mathbf{W}_{3}\mathbf{H}_{3},  \\
		\mathbf{G}_{1}=&\mathbf{X}_{1}-\mathbf{W}_{0}\mathbf{H}_{1}-\mathbf{W}_{1}\mathbf{H}_{0}-\mathbf{W}_{2}\mathbf{H}_{3}-\mathbf{W}_{3}\mathbf{H}_{2},   \\
		\mathbf{G}_{2}=&\mathbf{X}_{2}-\mathbf{W}_{0}\mathbf{H}_{2}-\mathbf{W}_{2}\mathbf{H}_{0}+\mathbf{W}_{1}\mathbf{H}_{3}+\mathbf{W}_{3}\mathbf{H}_{1},   \\
		\mathbf{G}_{3}=&\mathbf{X}_{3}-\mathbf{W}_{0}\mathbf{H}_{3}-\mathbf{W}_{3}\mathbf{H}_{0}-\mathbf{W}_{1}\mathbf{H}_{2}-\mathbf{W}_{2}\mathbf{H}_{1}.
	\end{split}
\end{equation*}
Thus, $f$ can be viewed as a function of the real-valued variables $\mathbf{W}_{p}\in\mathbb{R}^{M\times l}$ and $\mathbf{H}_{p}\in\mathbb{R}^{l\times N} (p=0,1,2,3)$. The gradients of $f$ with respect to $\mathbf{W}_{p}(p=0,1,2,3)$ are:
	\begin{equation*}\small
	\begin{split}
		\frac{\partial f}{\partial \mathbf{W}_{0}}=&-\mathbf{G}_{0}\mathbf{H}_{0}^{T}-\mathbf{G}_{1}\mathbf{H}_{1}^{T}-\mathbf{G}_{2}\mathbf{H}_{2}^{T}-\mathbf{G}_{3}\mathbf{H}_{3}^{T}, \\
		\frac{\partial f}{\partial \mathbf{W}_{1}}=&\mathbf{G}_{0}\mathbf{H}_{1}^{T}-\mathbf{G}_{1}\mathbf{H}_{0}^{T}+\mathbf{G}_{2}\mathbf{H}_{3}^{T}-\mathbf{G}_{3}\mathbf{H}_{2}^{T}, \\
		\frac{\partial f}{\partial \mathbf{W}_{2}}=&-\mathbf{G}_{0}\mathbf{H}_{2}^{T}-\mathbf{G}_{1}\mathbf{H}_{3}^{T}-\mathbf{G}_{2}\mathbf{H}_{0}^{T}-\mathbf{G}_{3}\mathbf{H}_{1}^{T}, \\
		\frac{\partial f}{\partial \mathbf{W}_{3}}=& \mathbf{G}_{0}\mathbf{H}_{3}^{T}-\mathbf{G}_{1}\mathbf{H}_{2}^{T}+\mathbf{G}_{2}\mathbf{H}_{1}^{T}-\mathbf{G}_{3}\mathbf{H}_{0}^{T}.
	\end{split}
\end{equation*}
Therefore, the steepest descent direction of $f$ with respect to $\mathbf{W}_{p}$ is $-\frac{\partial f}{\partial \mathbf{W}_{p}} (p=0,1,2,3)$. In other words, the steepest descent direction of $f$ with respect to $\ddot{\mathbf{W}}=\mathbf{W}_{0}+\mathbf{W}_{1}i+\mathbf{W}_{2}j+\mathbf{W}_{3}k$ can be written in a compact form: $-(\frac{\partial f}{\partial \mathbf{W}_{0}}+\frac{\partial f}{\partial \mathbf{W}_{1}}i+\frac{\partial f}{\partial \mathbf{W}_{2}}j+\frac{\partial f}{\partial \mathbf{W}_{3}}k)$. Based on this observation, to simplify the expressions and unify the update rules, we adopt the following notational convention for the combined quaternionic derivative term:
\begin{equation}\small\label{gwe}
	\begin{split}
		\nabla_{\ddot{\mathbf{W}}} f(\ddot{\mathbf{W}},\ddot{\mathbf{H}})&\overset{\Delta}{=}\frac{\partial f}{\partial \mathbf{W}_{0}}+\frac{\partial f}{\partial \mathbf{W}_{1}}i+\frac{\partial f}{\partial \mathbf{W}_{2}}j+\frac{\partial f}{\partial \mathbf{W}_{3}}k\\
		&=(\ddot{\mathbf{W}}\ddot{\mathbf{H}}-\ddot{\mathbf{X}})\ddot{\mathbf{H}}^{H},
	\end{split}
\end{equation}
where the second equality follows from direct computation. We call it `gradient' of $f$ with respect to $\ddot{\mathbf{W}}$ for convenience.
Similarly, we have:
	\begin{equation}\small\label{ghe}
	\begin{split}
		\nabla_{\ddot{\mathbf{H}}} f(\ddot{\mathbf{W}},\ddot{\mathbf{H}})&\overset{\Delta}{=}\frac{\partial f}{\partial \mathbf{H}_{0}}+\frac{\partial f}{\partial \mathbf{H}_{1}}i+\frac{\partial f}{\partial \mathbf{H}_{2}}j+\frac{\partial f}{\partial \mathbf{H}_{3}}k\\
		&=\ddot{\mathbf{W}}^{H}(\ddot{\mathbf{W}}\ddot{\mathbf{H}}-\ddot{\mathbf{X}}).
	\end{split}
\end{equation}
Note that these `gradient' definitions do not rely on RB calculus, but rather serve as a convenient representation of the steepest descent direction in the space of four real-valued matrix variables. This formulation allows the update rules to be expressed in a compact and unified form, which is particularly beneficial for the design and analysis of optimization algorithms.

\subsubsection{RB Projected Gradient Algorithm}
In order to tackle the optimization problem in (\ref{opp}), we consider the following RB alternating non-negative least squares (RB-ANNLS) problem, which involves fixing one RB matrix and optimizing the other in an alternating fashion:
\begin{equation}\small\label{anls}
	\left\{
	\begin{split}
		&  \ddot{\mathbf{W}}_{t+1}={\rm{arg\, min}}_{\ddot{\mathbf{W}}\in\mathbb{RB}^{M\times l}_{+}}\	\frac{1}{2}\|\ddot{\mathbf{X}}-\ddot{\mathbf{W}}\ddot{\mathbf{H}}_{t}\|_{F}^{2},\\
		& \ddot{\mathbf{H}}_{t+1}= {\rm{arg\, min}}_{\ddot{\mathbf{H}}\in\mathbb{RB}^{l\times N}_{+j}}\	\frac{1}{2}\|\ddot{\mathbf{X}}-\ddot{\mathbf{W}}_{t+1}\ddot{\mathbf{H}}\|_{F}^{2}.
	\end{split}
	\right.
\end{equation}
Let $\mathcal{P}_{\mathbb{RB}^{M\times N}_{+}}$ and $\mathcal{P}_{\mathbb{RB}^{M\times N}_{+j}}$ be two projections on $\mathbb{RB}^{M\times N}_{+}$ and $\mathbb{RB}^{M\times N}_{+j}$, which are defined as
\begin{equation*}\small
	\begin{split}
		\mathcal{P}_{\mathbb{RB}^{M\times N}_{+}}(\ddot{\mathbf{Q}})\overset{\Delta}{=}&\mathcal{P}_{\mathbb{R}^{M\times N}_{+}}(\mathbf{Q}_{0})+\mathcal{P}_{\mathbb{R}^{M\times N}_{+}}(\mathbf{Q}_{1})i\\
		&+\mathcal{P}_{\mathbb{R}^{M\times N}_{+}}(\mathbf{Q}_{2})j+\mathcal{P}_{\mathbb{R}^{M\times N}_{+}}(\mathbf{Q}_{3})k,\\
		\mathcal{P}_{\mathbb{RB}^{M\times N}_{+j}}(\ddot{\mathbf{Q}})\overset{\Delta}{=}&\mathcal{P}_{\mathbb{R}^{M\times N}_{+}}(\mathbf{Q}_{0})+\mathcal{P}_{\mathbb{R}^{M\times N}_{+}}(\mathbf{Q}_{2})j,
	\end{split}
\end{equation*}
where $\mathcal{P}_{\mathbb{R}^{M\times N}_{+}}(\mathbf{Q}_{l})\overset{\Delta}{=}\max(\mathbf{Q}_{l},0)$.

Obviously, for problem (\ref{anls}), the following standard RB projected gradient method can be applied:
\begin{equation}\small\label{RBpg}
	\left\{
	\begin{split}
		&  \ddot{\mathbf{W}}_{t+1}=\mathcal{P}_{\mathbb{RB}^{M\times l}_{+}}\big(\ddot{\mathbf{W}}_{t}-\alpha\nabla_{\ddot{\mathbf{W}}} f(\ddot{\mathbf{W}}_{t},\ddot{\mathbf{H}}_{t})\big),\\
		& \ddot{\mathbf{H}}_{t+1}= \mathcal{P}_{\mathbb{RB}^{l\times N}_{+j}}\big(\ddot{\mathbf{H}}_{t}-\beta\nabla_{\ddot{\mathbf{H}}} f(\ddot{\mathbf{W}}_{t+1},\ddot{\mathbf{H}}_{t})\big),
	\end{split}
	\right.
\end{equation}
where $\alpha$ and $\beta$ are the step sizes. To ensure the effective convergence of the algorithm, the selection of step size is necessary. In this work, we use the Armijo linear search method (\emph{i.e.}, inequalities (\ref{up_w}) and (\ref{up_h})) to determine the step size of each update. As a classic step size selection technique in projected gradient algorithms, this method was originally proposed in \cite{bertsekas1976goldstein} and has been applied in NMF \cite{lin2007projected} and QNQMF \cite{ke2023quasi}. Specifically, the algorithm procedure is shown in Algorithm \ref{tab_algorithm1}.

\begin{algorithm}\small
	\caption{RB projected gradient method for RB-ANNLS problem (\ref{anls}). }
	\hrule
	\label{tab_algorithm1}
	\begin{algorithmic}[1]
		\REQUIRE Given an RB matrix $\ddot{\mathbf{X}}\in\mathbb{RB}^{M\times N}_{+}$, $l<\min(M,N)$, $tol=10^{-4}$, $\mu=0.1$, $\sigma=0.001$, and maximum number of iterations $I$.
		\STATE \textbf{Initialize} $\ddot{\mathbf{W}}_{0}\in\mathbb{RB}^{M\times l}_{+}$ and $\ddot{\mathbf{H}}_{0}\in\mathbb{RB}^{l\times N}_{+j}$, are randomly initialized; $t=0$.
		\STATE \textbf{Repeat}
		\STATE \% Update $\ddot{\mathbf{W}}$
		\STATE Compute $\nabla_{\ddot{\mathbf{W}}} f(\ddot{\mathbf{W}}_{t},\ddot{\mathbf{H}}_{t})=(\ddot{\mathbf{W}}_{t}\ddot{\mathbf{H}}_{t}-\ddot{\mathbf{X}})\ddot{\mathbf{H}}_{t}^{H}$;
		\STATE Update $\ddot{\mathbf{W}}_{t+1}=\mathcal{P}_{\mathbb{RB}^{M\times l}_{+}}\big(\ddot{\mathbf{W}}_{t}-\alpha_{t}\nabla_{\ddot{\mathbf{W}}} f(\ddot{\mathbf{W}}_{t},\ddot{\mathbf{H}}_{t})\big)$, where $\alpha_{t}=\mu^{d_{t}}$, and $d_{t}$ is the first non-negative integer $d$ for which
		\begin{equation}\small\label{up_w}
			\begin{split}
			&f(\ddot{\mathbf{W}}_{t+1},\ddot{\mathbf{H}}_{t})-f(\ddot{\mathbf{W}}_{t},\ddot{\mathbf{H}}_{t})\\
			&\leq\sigma {\rm{Re}}\left(\langle \nabla_{\ddot{\mathbf{W}}} f(\ddot{\mathbf{W}}_{t},\ddot{\mathbf{H}}_{t}), \ddot{\mathbf{W}}_{t+1}-\ddot{\mathbf{W}}_{t} \rangle\right).
			\end{split}
		\end{equation}
		\STATE \% Update $\ddot{\mathbf{H}}$
		\STATE Compute $\nabla_{\ddot{\mathbf{H}}} f(\ddot{\mathbf{W}}_{t+1},\ddot{\mathbf{H}}_{t})=\ddot{\mathbf{W}}_{t+1}^{H}(\ddot{\mathbf{W}}_{t+1}\ddot{\mathbf{H}}_{t}-\ddot{\mathbf{X}})$;
		\STATE Update $\ddot{\mathbf{H}}_{t+1}=\mathcal{P}_{\mathbb{RB}^{l\times N}_{+j}}\big(\ddot{\mathbf{H}}_{t}-\beta_{t}\nabla_{\ddot{\mathbf{H}}} f(\ddot{\mathbf{W}}_{t+1},\ddot{\mathbf{H}}_{t})\big)$, where $\beta_{t}=\mu^{s_{t}}$, and $s_{t}$ is the first non-negative integer $s$ for which
		\begin{equation}\small\label{up_h}
			\begin{split}
			&f(\ddot{\mathbf{W}}_{t+1},\ddot{\mathbf{H}}_{t+1})-f(\ddot{\mathbf{W}}_{t+1},\ddot{\mathbf{H}}_{t})\\
			&\leq\sigma {\rm{Re}}\left(\langle \nabla_{\ddot{\mathbf{H}}} f(\ddot{\mathbf{W}}_{t+1},\ddot{\mathbf{H}}_{t}), \ddot{\mathbf{H}}_{t+1}-\ddot{\mathbf{H}}_{t} \rangle\right).
	    	\end{split}
		\end{equation}
		\STATE  $t\longleftarrow t+1$.
		\STATE \textbf{Until} $\frac{\|\ddot{\mathbf{W}}_{t}\ddot{\mathbf{H}}_{t}-\ddot{\mathbf{W}}_{t-1}\ddot{\mathbf{H}}_{t-1}\|_{F}}{\|\ddot{\mathbf{W}}_{t-1}\ddot{\mathbf{H}}_{t-1}\|_{F}}<tol$ or $t-1>I$.
		\ENSURE  $\ddot{\mathbf{W}}_{t}\in\mathbb{RB}^{M\times l}_{+}$ and $\ddot{\mathbf{H}}_{t}\in\mathbb{RB}^{l\times N}_{+j}$.
	\end{algorithmic}
\end{algorithm}

\begin{remark}\label{remark3}
	The computationally expensive step in Algorithm~1 is to determine $\alpha_t$ and $\beta_t$. To minimize checks, we initialize them using the previous values ($\alpha_{t-1}$, $\beta_{t-1}$) and adjust to find the largest valid step sizes satisfying (\ref{up_w}) and (\ref{up_h}). This leverages the similarity between consecutive step sizes \cite{lin2007projected,lin1999more} and often enables faster convergence by projecting variables to their bounds more efficiently. We refer to the algorithm using this strategy as the \textbf{RB} \textbf{I}mproved \textbf{P}rojected \textbf{G}radient (\textbf{RBIPG}) algorithm, which is the algorithm ultimately adopted in this study.
\end{remark}

\subsubsection{Convergence analysis}\label{convergence}
To theoretically characterize the stationary points to which Algorithm \ref{tab_algorithm1} may converge (as guaranteed by Theorem \ref{theorem_con2}), we first derive the necessary optimality conditions for problem (\ref{opp}) using the Karush-Kuhn-Tucker (KKT) framework in the following property.
\begin{property}\label{property1}
If	$(\ddot{\mathbf{W}}^{\#},\ddot{\mathbf{H}}^{\#})$ is a stationary point of the optimization problem (\ref{opp}), then it satisfies:
	\begin{equation}\small\label{kkt}
		\left\{
		\begin{split}
			&\ddot{\mathbf{W}}^{\#}\in\mathbb{RB}^{M\times l}_{+},\ \ddot{\mathbf{H}}^{\#}\in\mathbb{RB}^{l\times N}_{+j};\\
			& \nabla_{\ddot{\mathbf{W}}} f(\ddot{\mathbf{W}}^{\#},\ddot{\mathbf{H}}^{\#})\in\mathbb{RB}^{M\times l}_{+},\ \nabla_{\ddot{\mathbf{H}}} f(\ddot{\mathbf{W}}^{\#},\ddot{\mathbf{H}}^{\#})\in\mathbb{RB}^{l\times N}_{+j};\\
			&{\rm{Re}}(\ddot{\mathbf{W}}^{\#})\otimes{\rm{Re}}(\nabla_{\ddot{\mathbf{W}}} f(\ddot{\mathbf{W}}^{\#},\ddot{\mathbf{H}}^{\#}))=0,\\ 
			&{\rm{Im}}_{\eta}(\ddot{\mathbf{W}}^{\#})\otimes{\rm{Im}}_{\eta}(\nabla_{\ddot{\mathbf{W}}} f(\ddot{\mathbf{W}}^{\#},\ddot{\mathbf{H}}^{\#}))=0,\ \eta=i,j,k;\\
			&{\rm{Re}}(\ddot{\mathbf{H}}^{\#})\otimes{\rm{Re}}(\nabla_{\ddot{\mathbf{H}}} f(\ddot{\mathbf{W}}^{\#},\ddot{\mathbf{H}}^{\#}))=0,\\
			&{\rm{Im}}_{j}(\ddot{\mathbf{H}}^{\#})\otimes{\rm{Im}}_{j}(\nabla_{\ddot{\mathbf{H}}} f(\ddot{\mathbf{W}}^{\#},\ddot{\mathbf{H}}^{\#}))=0.
		\end{split}
		\right.
	\end{equation}
\end{property}
The proof of Property \ref{property1} can be found in Section I of the Supplementary Materials.

\begin{corollary}\label{corollary1}
Conditions in (\ref{kkt}) are equivalent to the following conditions:
	\begin{equation}\small\label{kktcorol}
		\left\{
		\begin{split}
			&{\rm{Re}}\left(\langle \nabla_{\ddot{\mathbf{W}}} f(\ddot{\mathbf{W}}^{\#},\ddot{\mathbf{H}}^{\#}), \ddot{\mathbf{M}}-\ddot{\mathbf{W}}^{\#} \rangle\!\right)\!\geq 0, \  \forall \ \! \ddot{\mathbf{M}}\in\mathbb{RB}^{M\times l}_{+}, \\
			&{\rm{Re}}\left(\langle \nabla_{\ddot{\mathbf{H}}} f(\ddot{\mathbf{W}}^{\#},\ddot{\mathbf{H}}^{\#}), \ddot{\mathbf{N}}-\ddot{\mathbf{H}}^{\#} \rangle\right)\geq 0, \ \forall \ \ddot{\mathbf{N}}\in\mathbb{RB}^{l\times N}_{+j}.
		\end{split}
		\right.
	\end{equation}
\end{corollary}
The proof of Corollary \ref{corollary1} can be found in Section II of the Supplementary Materials.

For analytical convenience, we focus on the convergence of RBPG in Algorithm \ref{tab_algorithm1}. RBIPG, introduced in Remark \ref{remark3}, is a special case of RBPG with specific step sizes; thus, its convergence properties align with those of RBPG.

\begin{theorem}\label{theorem_con1}
	The sequence of objective function values $\{f(\ddot{\mathbf{W}}_{t},\ddot{\mathbf{H}}_{t})\}_{t=1}^{\infty}$ generated by RBPG in Algorithm \ref{tab_algorithm1} is nonincreasing.
\end{theorem}
The proof of Theorem \ref{theorem_con1} can be found in Section III of the Supplementary Materials.

\begin{theorem}\label{theorem_con2}
	Suppose $\{f(\ddot{\mathbf{W}}_{t},\ddot{\mathbf{H}}_{t})\}_{t=1}^{\infty}$ is the sequence generated by RBPG in Algorithm \ref{tab_algorithm1}. If
	\begin{equation*}\small
		\lim_{t \to +\infty}\ddot{\mathbf{W}}_{t}=\ddot{\mathbf{W}}^{\#} \quad \text{and} \quad 	\lim_{t \to +\infty}\ddot{\mathbf{H}}_{t}=\ddot{\mathbf{H}}^{\#},
	\end{equation*}
	then $(\ddot{\mathbf{W}}^{\#},\ddot{\mathbf{H}}^{\#})$ is a stationary point of (\ref{opp}).
\end{theorem}
The proof of Theorem \ref{theorem_con2} can be found in Section IV of the Supplementary Materials.

\section{Numerical Experiments}\label{experiments}

Extensive numerical experiments on color face recognition are conducted in this section. All experiments are run in MATLAB $2014b$ under Windows $10$ on a laptop with a $1.60$GHz CPU and $8$GB of memory.

\subsection{Applications in Color Face Recognition}\label{acfr}

In this subsection, we give the detailed process of performing color face recognition using the proposed NRBMF model.

\textbf{$1)$ Preparing the training and testing data sets.} All color faces are represented as an RB matrix in the form of (\ref{face_re}). Assuming we have $K$ training samples $\ddot{\mathbf{T}}_{k}\in\mathbb{RB}^{M\times N}_{+}$, $k=1,2,\ldots,K$, and $S$ testing samples $\ddot{\mathbf{P}}_{s}\in\mathbb{RB}^{M\times N}_{+}$, $s=1,2,\ldots,S$. Let $\ddot{\mathbf{X}}=[\ddot{\mathbf{t}}_{1},\ddot{\mathbf{t}}_{2},\ldots,\ddot{\mathbf{t}}_{K}]\in\mathbb{RB}^{MN\times K}_{+}$, $\ddot{\mathbf{G}}=[\ddot{\mathbf{p}}_{1},\ddot{\mathbf{p}}_{2},\ldots,\ddot{\mathbf{p}}_{S}]\in\mathbb{RB}^{MN\times S}_{+}$, where $\ddot{\mathbf{t}}_{k}={\rm{vec}}(\ddot{\mathbf{T}}_{k})\in\mathbb{RB}^{MN\times 1}_{+}$ for $k=1,2,\ldots,K$ and $\ddot{\mathbf{p}}_{s}={\rm{vec}}(\ddot{\mathbf{P}}_{s})\in\mathbb{RB}^{MN\times 1}_{+}$ for $s=1,2,\ldots,S$. 

\textbf{$2)$ Obtaining the basis matrix and encoding coefficients.} For the given $\ddot{\mathbf{X}}\in\mathbb{RB}^{MN\times K}_{+}$ and $l$, we apply RBIPG algorithm to obtain the basis matrix $\ddot{\mathbf{W}}\in\mathbb{RB}^{MN\times l}_{+}$. From  $\ddot{\mathbf{X}}=\ddot{\mathbf{W}}\ddot{\mathbf{H}}$, the encodings  $\ddot{\mathbf{h}}_{k}^{(train)}$ of each training face $\ddot{\mathbf{t}}_{k}$, is given by $ \ddot{\mathbf{h}}_{k}^{(train)}=(\ddot{\mathbf{W}}^{H}\ddot{\mathbf{W}})^{-1}\ddot{\mathbf{W}}^{H}\ddot{\mathbf{t}}_{k}$ \cite{el2022linear}.

\textbf{$3)$ Color face recognition.} For a given test color face $\ddot{\mathbf{p}}_{s}\in\mathbb{RB}^{MN\times 1}_{+}$, based on the basis matrix $\ddot{\mathbf{W}}\in\mathbb{RB}^{MN\times l}_{+}$, the corresponding encodings $\ddot{\mathbf{h}}_{s}^{test}\in\mathbb{RB}^{l\times 1}$
can be obtained by $\ddot{\mathbf{h}}_{s}^{test}=(\ddot{\mathbf{W}}^{H}\ddot{\mathbf{W}})^{-1}\ddot{\mathbf{W}}^{H}\ddot{\mathbf{p}}_{s}$.

The following cosine similarity measure is utilized to evaluate the similarity between a given test sample and the training samples: 
\begin{equation}\small\label{c_measure}
	d_{k}=\frac{{\rm{Re}}(\langle\ddot{\mathbf{h}}_{k}^{(train)},\ddot{\mathbf{h}}_{s}^{test}\rangle)}{\|\ddot{\mathbf{h}}_{k}^{(train)}\|_{F}\|\ddot{\mathbf{h}}_{s}^{test}\|_{F}},\quad  k=1,2,\ldots,K.
\end{equation}
Denote $d_{k^{\bullet}}=\max\{d_{1},d_{2},\ldots,d_{K}\}$, then the test color face $\ddot{\mathbf{p}}_{s}$ is considered to belong to the subject that the training sample $\ddot{\mathbf{t}}_{k^{\bullet}}$ belongs to.

\begin{remark} If
 $\max\{{\rm{cond}}(\mathbf{M}_{1}),{\rm{cond}}(\mathbf{M}_{2})\}>10^{15}$, the RB gradient descent method is used in steps $2)$ and $3)$, to solve the RB least-square problems $	\mathop{{\rm{min}}}\limits_{\ddot{\mathbf{h}}\in\mathbb{RB}^{l\times 1}}\ \|\ddot{\mathbf{t}}_{k}-\ddot{\mathbf{W}}\ddot{\mathbf{h}}\|_{F}^{2}$ and $	\mathop{{\rm{min}}}\limits_{\ddot{\mathbf{h}}\in\mathbb{RB}^{l\times 1}}\ \|\ddot{\mathbf{p}}_{s}-\ddot{\mathbf{W}}\ddot{\mathbf{h}}\|_{F}^{2}$
to obtain $\ddot{\mathbf{h}}_{k}^{(train)}$ and $\ddot{\mathbf{h}}_{s}^{test}$ respectively.
\end{remark}

\subsection{Different Methods for Comparison}

This subsection presents comprehensive experiments validating the effectiveness and superiority of the proposed NRBMF model for color face recognition. We compare with the following several methods:
\begin{itemize}
	\item \textbf{RBIPG-full}: The RBIPG algorithm for NRBMF. The term `full' means representing the color face as shown in (\ref{face_re}), where the real component of $\ddot{\mathbf{X}}$ is $\mathbf{X}_{av}$.
	\item \textbf{RBIPG-pure}: The term `pure' means that the real component of $\ddot{\mathbf{X}}$ is a zero matrix. 
	\item \textbf{QIPG-full} \cite{ke2023quasi}: The quaternion improved projected gradient (QIPG) algorithm for QNQMF. For a fair comparison, in this scenario, we encode $\mathbf{X}_{av}$ as the real component of the quaternion matrix $\dot{\mathbf{X}}$.
	\item \textbf{QIPG-pure} \cite{ke2023quasi}: The same as its original setting, \emph{i.e.}, the real component of quaternion matrix $\dot{\mathbf{X}}$ is zero.
	\item \textbf{RIPG-full} \cite{lin2007projected}: The real improved projected gradient (RIPG) algorithm for NMF. In this setting, we perform NMF on $\mathbf{X}_{av}$, $\mathbf{X}_{R}$, $\mathbf{X}_{G}$, and $\mathbf{X}_{B}$ respectively, \emph{i.e.}, $\mathbf{X}_{av}=\mathbf{W}_{av}\mathbf{H}_{av}, \mathbf{X}_{R}=\mathbf{W}_{R}\mathbf{H}_{R}, \mathbf{X}_{G}=\mathbf{W}_{G}\mathbf{H}_{G}, \mathbf{X}_{B}=\mathbf{W}_{B}\mathbf{H}_{B}$.
	The face recognition process is also conducted in each component. When calculating the cosine similarity measure, (\ref{c_measure}) is replaced by the sum of four components, that is, the setting method in the \cite{ke2023quasi} is used.
	\item \textbf{RIPG-pure} \cite{lin2007projected}: Same as RIPG-full, except only $\mathbf{X}_{R}$, $\mathbf{X}_{G}$, and $\mathbf{X}_{B}$ are considered.
	\item \textbf{RIPG-full$^{v}$} \cite{lin2007projected}: Same as RIPG-full, except the four-channel color face image is vectorized for representation.
	\item \textbf{RIPG-pure$^{v}$} \cite{lin2007projected}: Same as RIPG-pure, except the three-channel color face image is vectorized for representation.
	\item \textbf{QPCA-full} \cite{liu2022quaternion}: The quaternion principal component analysis (QPCA) method for color face recognition, which encodes $\mathbf{X}_{av}$ as the real component of the quaternion matrix $\dot{\mathbf{X}}$. 
	\item \textbf{QPCA-pure} \cite{liu2022quaternion}: Follows the original setting, where the real component of the quaternion matrix $\dot{\mathbf{X}}$ is set to zero.
\end{itemize}

\begin{remark}
	The purpose of choosing the above comparison methods is as follows: Comparing RBIPG with QIPG mainly aims to demonstrate the superiority of our model, NRBMF, after overcoming the model drawbacks of QNQMF (as discussed in Subsection \ref{rw}). Comparing RBIPG with RIPG aims to highlight the superiority of the RB matrix over the real matrix in representing color images, as it treats the color channels as an integrated whole, fully utilizing and preserving the potential relationships between color channels. While QPCA is a representative quaternion-based color face recognition method in recent years.
\end{remark}

\subsubsection{Color Face Recognition for AR Database \cite{martinez1998ar}}\label{ar_exper}
\begin{figure}[htbp]
	\centering
	\subfigure[Samples of one subject from AR database.]{
		\includegraphics[width=7.8cm,height=1.3cm]{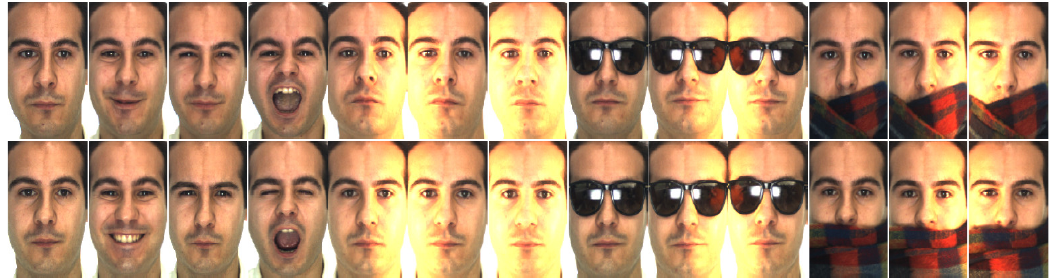}
	}\\
	\subfigure[Samples of one subject from KDEF database.]{
		\includegraphics[width=7.8cm,height=1.3cm]{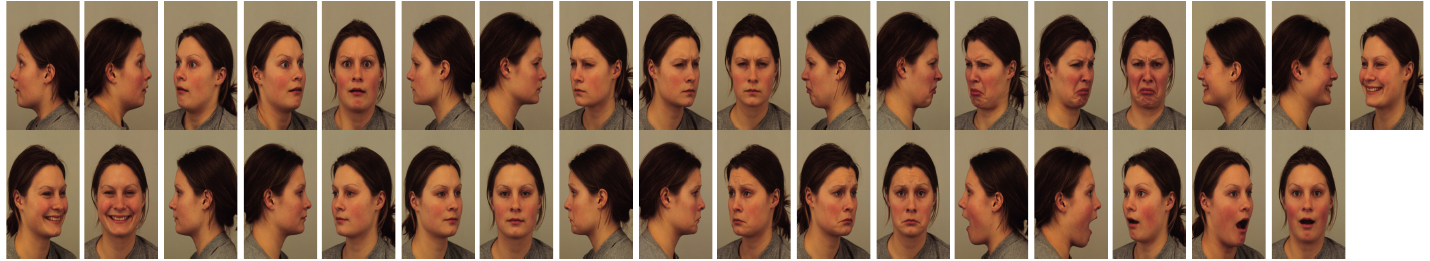}
	}
	\caption{Some samples of the used color face database.}
	\label{fig_face_ar}
\end{figure}

This experiment employs a standard AR database subset comprising $100$ subjects ($50$ male/$50$ female) \cite{wright2008robust}, with each subject providing $26$ images across two sessions under varying conditions (illumination, expressions, and occlusions including sunglasses/scarves). Figure \ref{fig_face_ar}(a) displays samples of one subject.
We set up the experiments in the following two scenarios:
\begin{enumerate}
	\item [(a)] This study uses $14$ non-occluded color face images per subject, randomly split into $7$ training and $7$ test images, all resized to $80 \times 60$  pixels.
	\item [(b)] The experiment retains all occluded faces ($26$ images per subject), using $13$ Session 1 images for training and $13$ Session 2 images for testing, with images resized to $80 \times 60$  pixels.
\end{enumerate}

\begin{figure*}[htbp]
	\centering
	\subfigure[The `full' case for different methods, $l=50$.]{
		\includegraphics[width=7cm,height=4.5cm]{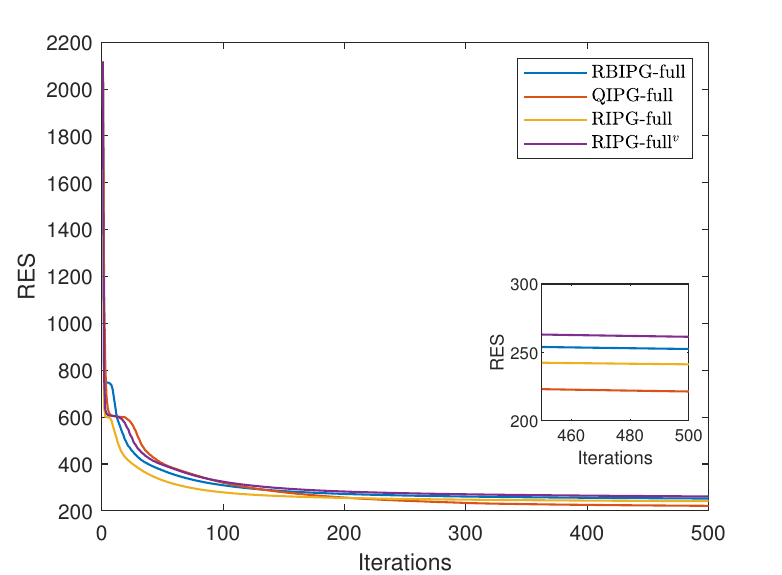}
	}
	\subfigure[The `pure' case for different methods, $l=50$.]{
		\includegraphics[width=7cm,height=4.5cm]{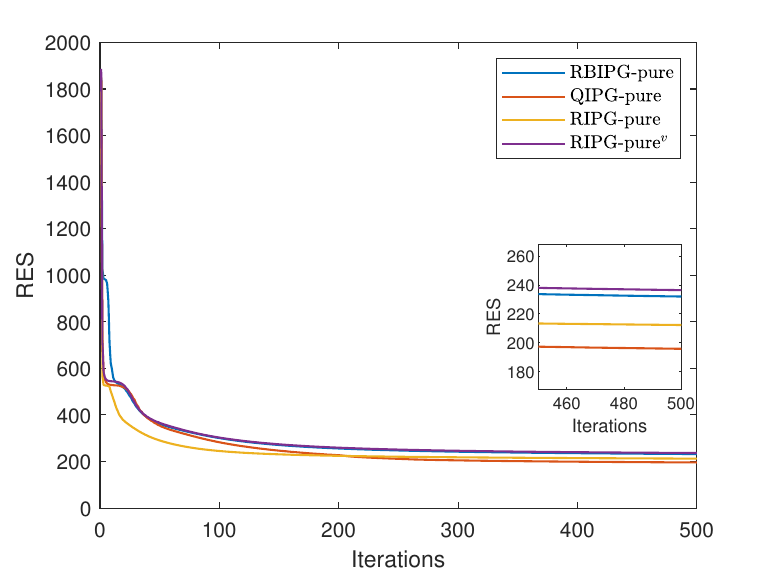}
	}
	\caption{RES versus iterations for different methods under experiment \ref{ar_exper}, Scenario (a), $l=50$.}
	\label{rse}
\end{figure*}

\begin{figure*}[htbp]
	\centering
	\includegraphics[width=15cm,height=2.8cm]{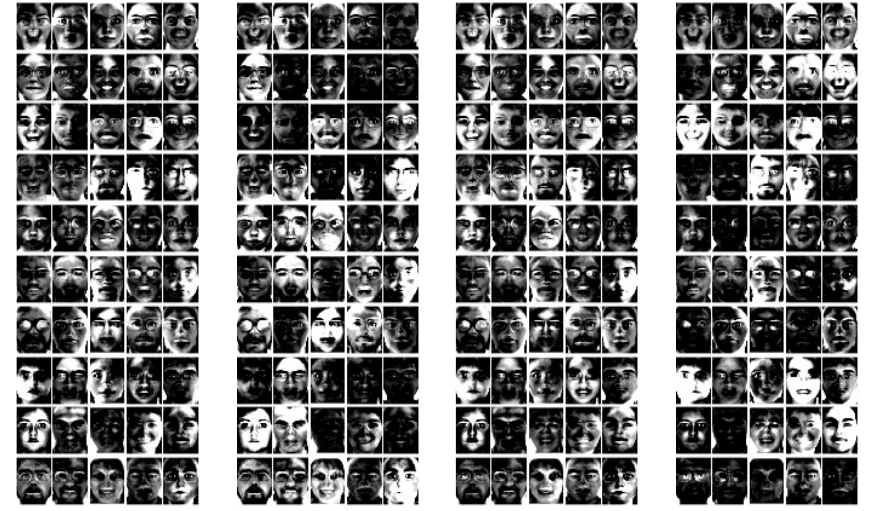}
\subfigure{(a) Basis images computed by RBIPG-full; From left to right are the ${\rm{Re}}(\ddot{\mathbf{W}})$, ${\rm{Im}}_{i}(\ddot{\mathbf{W}})$, ${\rm{Im}}_{j}(\ddot{\mathbf{W}})$, and ${\rm{Im}}_{k}(\ddot{\mathbf{W}})$.}\\
	\includegraphics[width=15cm,height=2.8cm]{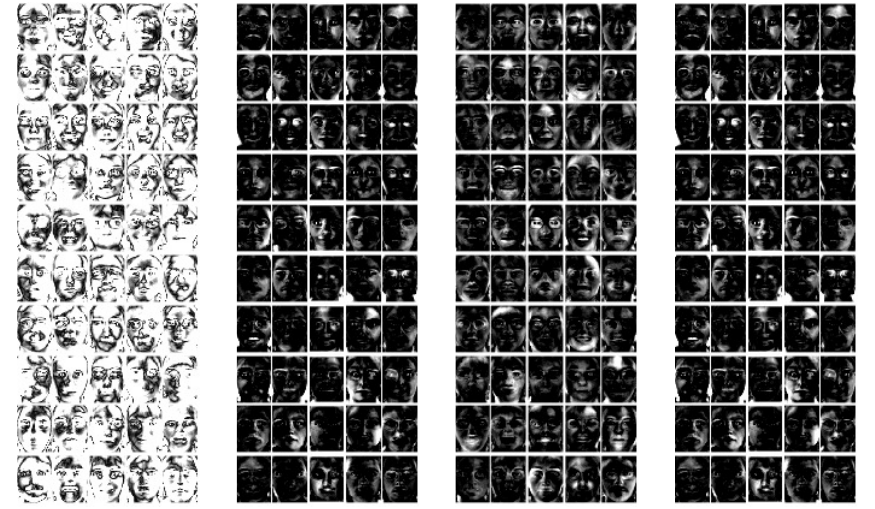}
\subfigure{(b) Basis images computed by QIPG-full; From left to right are the ${\rm{abs}}({\rm{Re}}(\dot{\mathbf{W}}))$, ${\rm{Im}}_{\mathbf{i}}(\dot{\mathbf{W}})$, ${\rm{Im}}_{\mathbf{j}}(\dot{\mathbf{W}})$, and  ${\rm{Im}}_{\mathbf{k}}(\dot{\mathbf{W}})$. }\\
	\includegraphics[width=15cm,height=2.8cm]{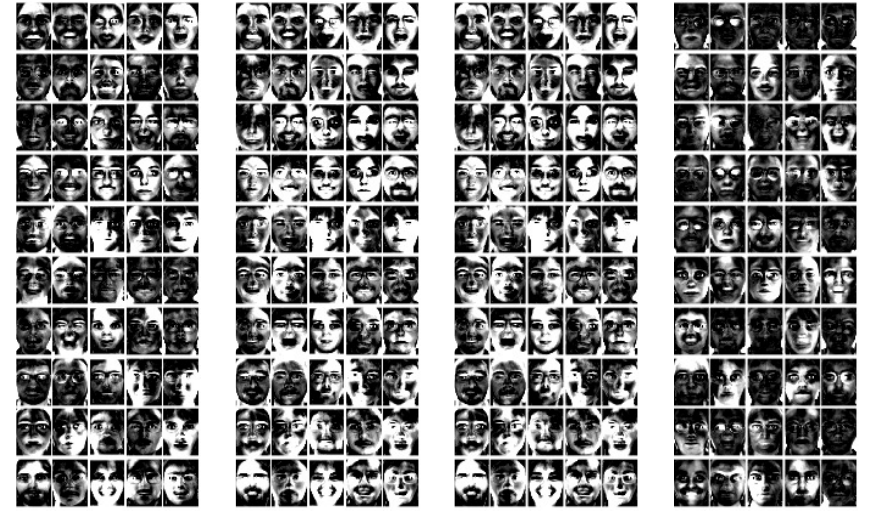}
\subfigure{(c) Basis images computed by RIPG-full; From left to right are the $\mathbf{W}_{av}$, $\mathbf{W}_{R}$, $\mathbf{W}_{G}$, and $\mathbf{W}_{B}$. }
\vspace{-0.2cm}
\caption{Basis images of different methods with the `full' case under
Scenario (a) of experiment \ref{ar_exper} with $l=50$.}
\label{bi_RB_Q_ADMM1}


\includegraphics[width=15cm,height=2.8cm]{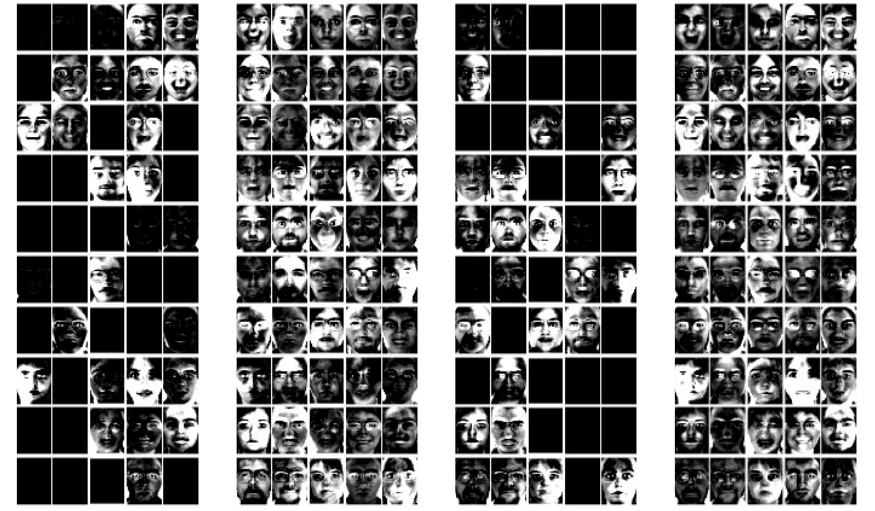}
\subfigure{(a) Basis images computed by RBIPG-pure; From left to right are the  ${\rm{Re}}(\ddot{\mathbf{W}})$, ${\rm{Im}}_{i}(\ddot{\mathbf{W}})$, ${\rm{Im}}_{j}(\ddot{\mathbf{W}})$, and ${\rm{Im}}_{k}(\ddot{\mathbf{W}})$.}\\
\includegraphics[width=15cm,height=2.8cm]{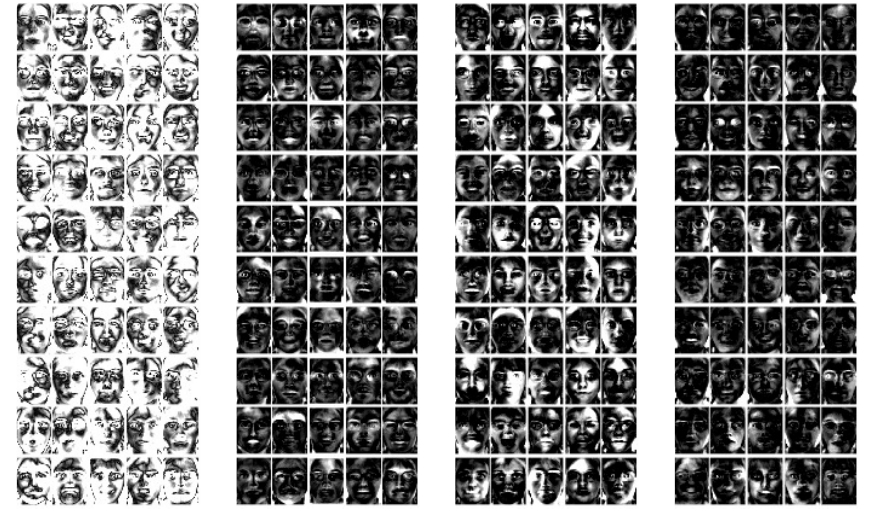}
\subfigure{(b) Basis images computed by QIPG-pure; From left to right are the ${\rm{abs}}({\rm{Re}}(\dot{\mathbf{W}}))$, ${\rm{Im}}_{\mathbf{i}}(\dot{\mathbf{W}})$, ${\rm{Im}}_{\mathbf{j}}(\dot{\mathbf{W}})$, and ${\rm{Im}}_{\mathbf{k}}(\dot{\mathbf{W}})$. }\\
\includegraphics[width=15cm,height=2.8cm]{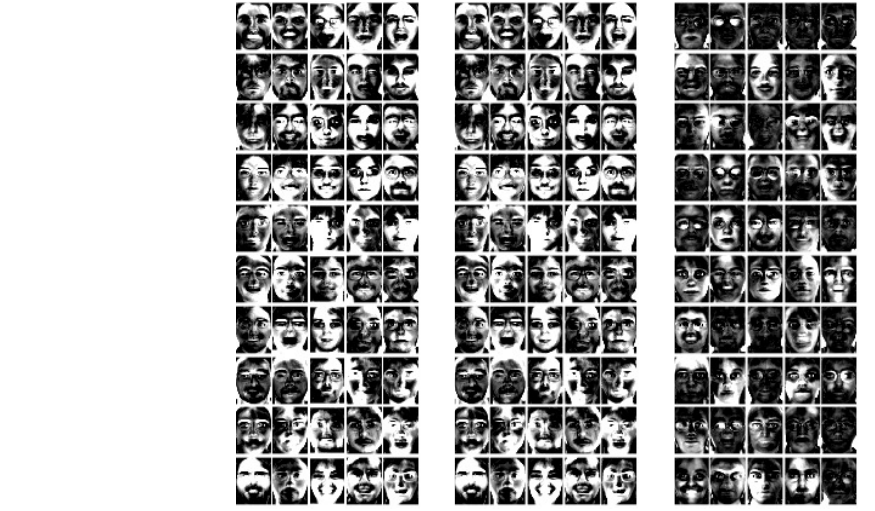}
\subfigure{(c) Basis images computed by RIPG-pure; From left to right are the  $\mathbf{W}_{R}$, $\mathbf{W}_{G}$, and $\mathbf{W}_{B}$.}
\vspace{-0.2cm}
\caption{Basis images of different methods with the `pure' case under
		Scenario (a) of experiment \ref{ar_exper} with $l=50$.}
		\label{bi_RB_Q_ADMM2}
\end{figure*}

\begin{figure*}[htbp]
	\centering
	\subfigure[Encoding coefficients computed by RBIPG-full; From left to right are the $10$th column (first row) and the $100$th column (second row) of ${\rm{Re}}(\ddot{\mathbf{H}})$ and   ${\rm{Im}}_{j}(\ddot{\mathbf{H}})$.]{
		\includegraphics[width=15cm,height=3.5cm]{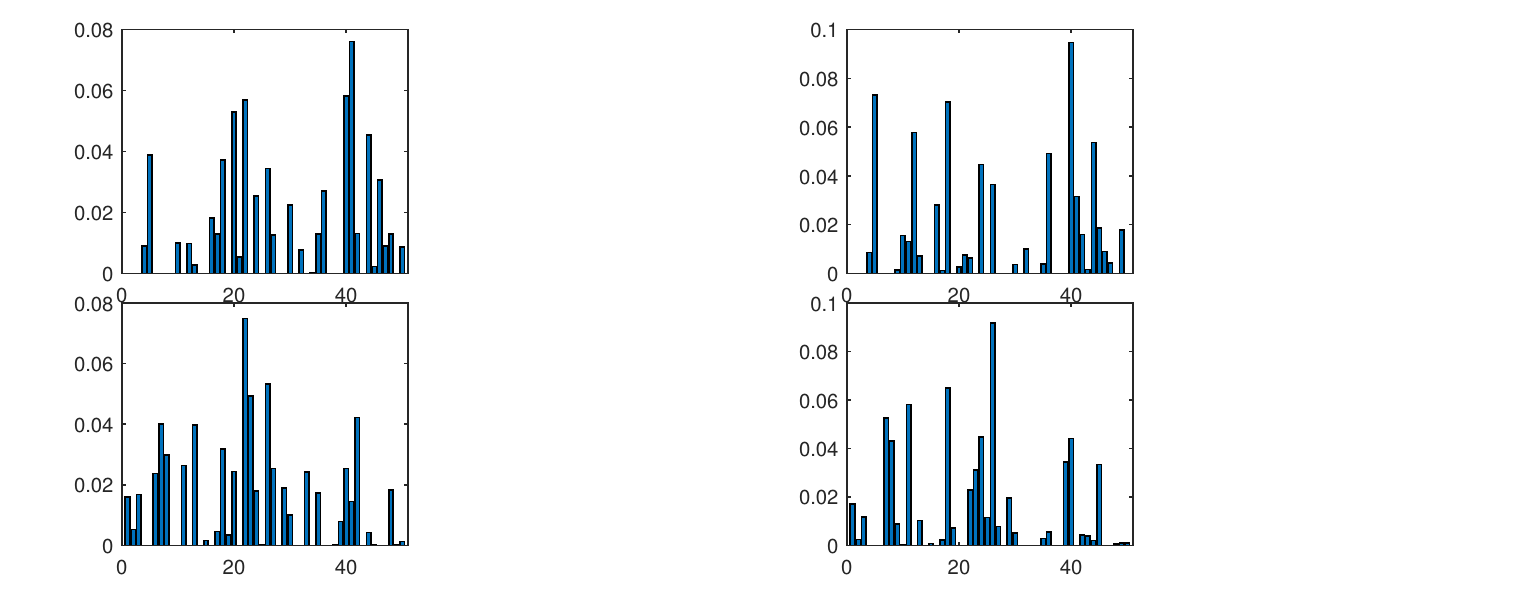}
	}
	\subfigure[Encoding coefficients computed by QIPG-full; From left to right are the $10$th column (first row) and the $100$th column (second row) of ${\rm{Re}}(\dot{\mathbf{H}})$,    ${\rm{Im}}_{\mathbf{i}}(\dot{\mathbf{H}})$, ${\rm{Im}}_{\mathbf{j}}(\dot{\mathbf{H}})$, and ${\rm{Im}}_{\mathbf{k}}(\dot{\mathbf{H}})$. ]{
		\includegraphics[width=15cm,height=3.5cm]{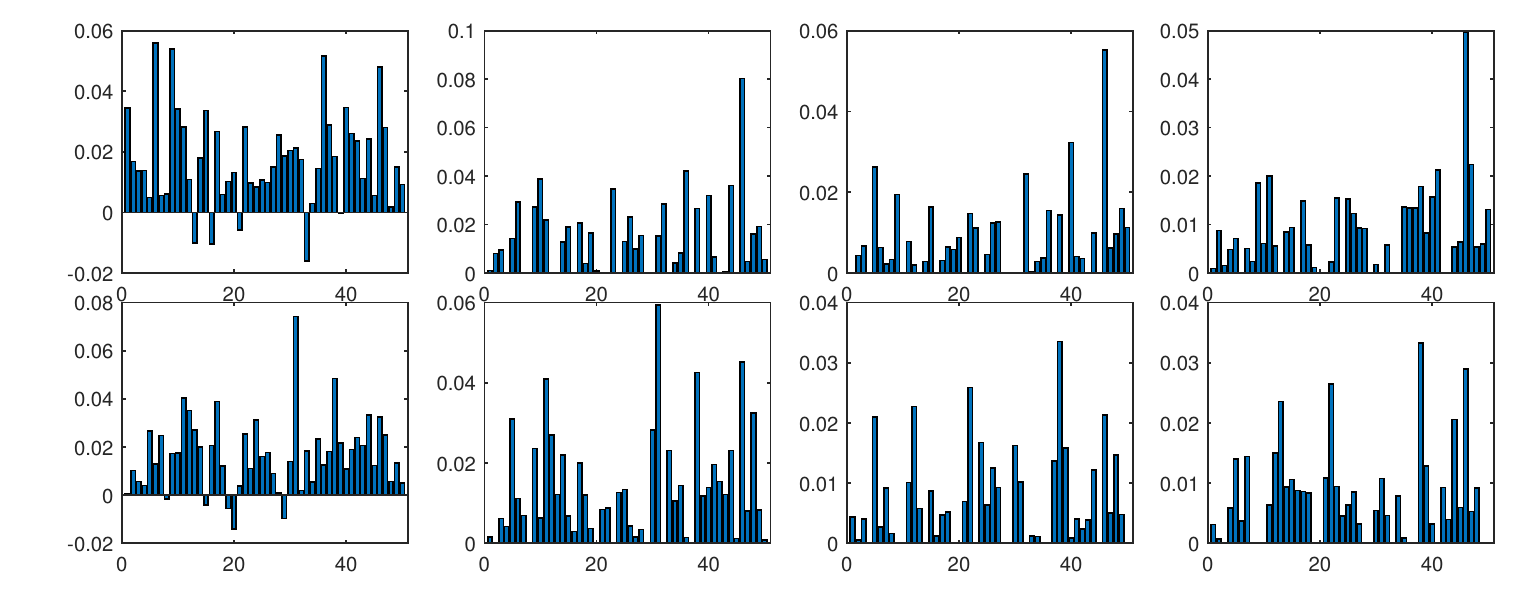}
	}
	\subfigure[Encoding coefficients computed by RIPG-full; From left to right are the $10$th column (first row) and the $100$th column (second row) of $\mathbf{H}_{av}$, $\mathbf{H}_{R}$, $\mathbf{H}_{G}$, and $\mathbf{H}_{B}$.]{
		\includegraphics[width=15cm,height=3.5cm]{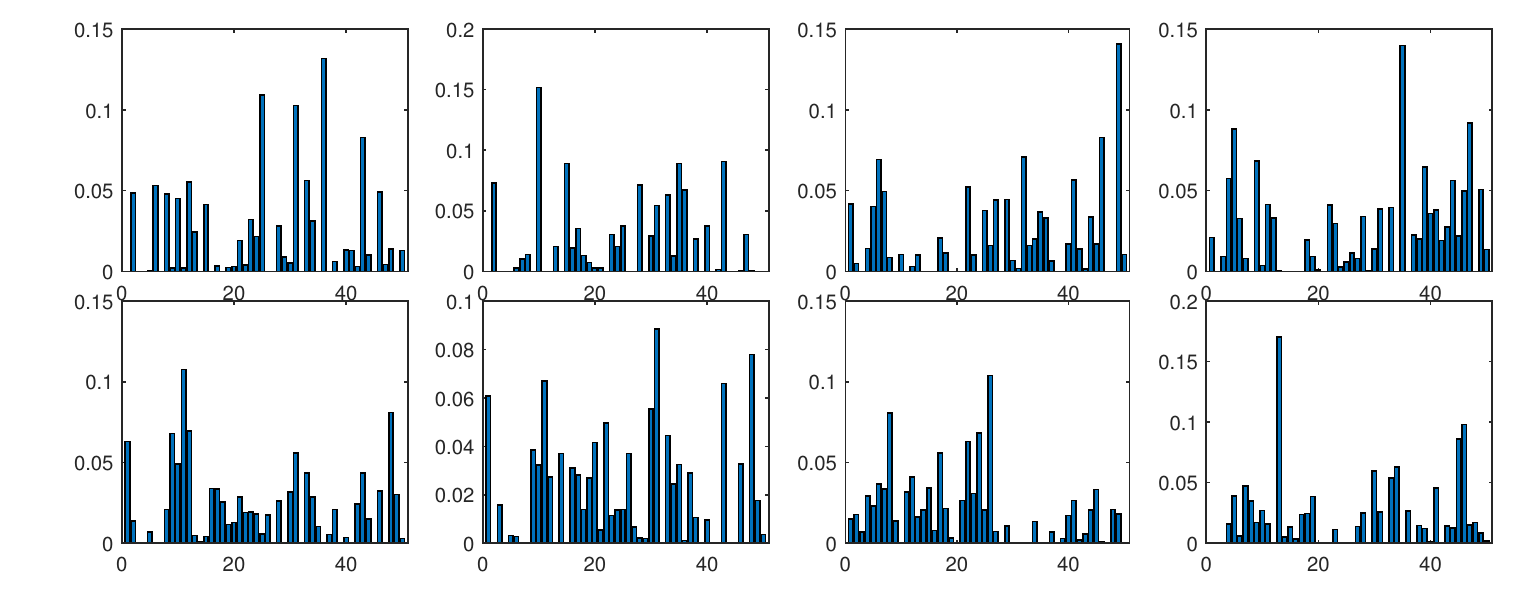}
	}
	\caption{The encoding coefficients of RBIPG-full (a),  QIPG-full (b), and RIPG-full (c) under
		Scenario (a) of experiment \ref{ar_exper} with $l=50$.
	}
	\label{bi_RB_Q_ADMM3}
\end{figure*}

\begin{figure*}[htbp]
	\centering
	\subfigure[The `full' case for Scenario (a)]{
		\includegraphics[width=7cm,height=4cm]{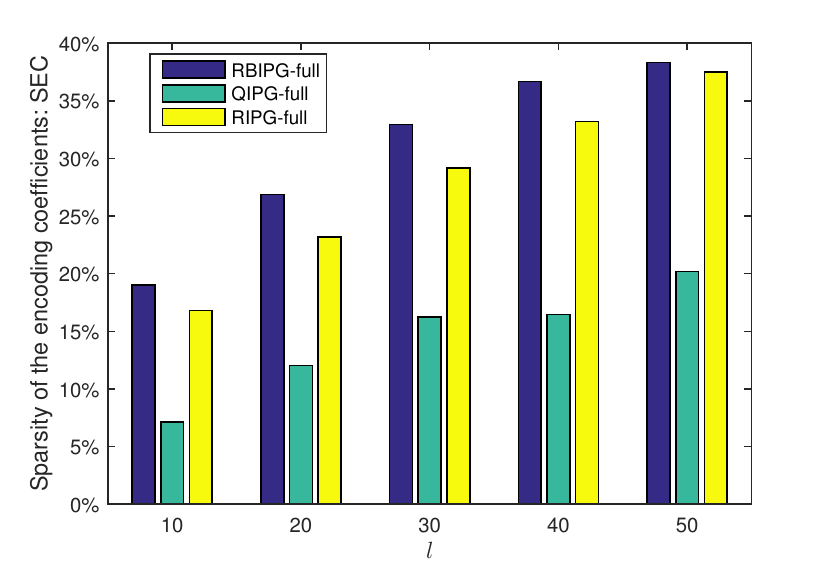}
	}
	\subfigure[The `full' case for Scenario (b)]{
		\includegraphics[width=7cm,height=4cm]{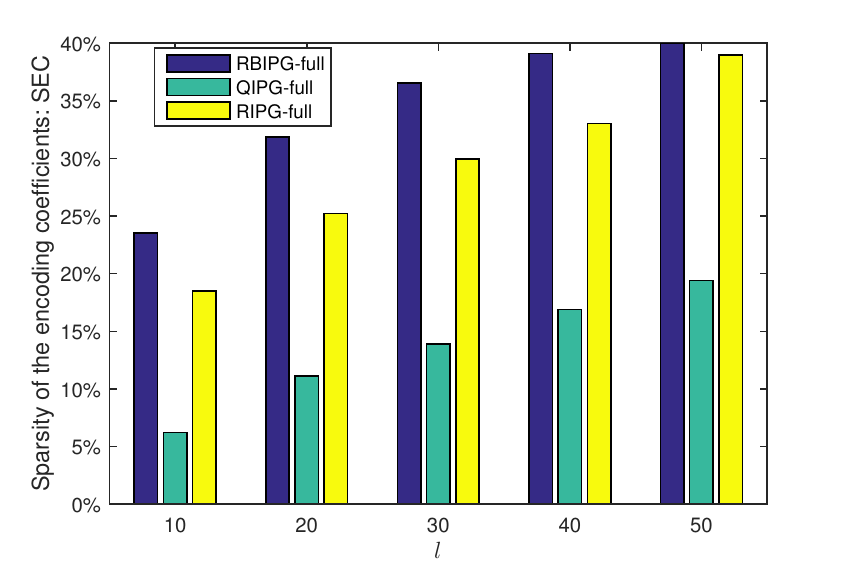}
	}\\
	\subfigure[The `full' case for Scenario (c)]{
		\includegraphics[width=7cm,height=4cm]{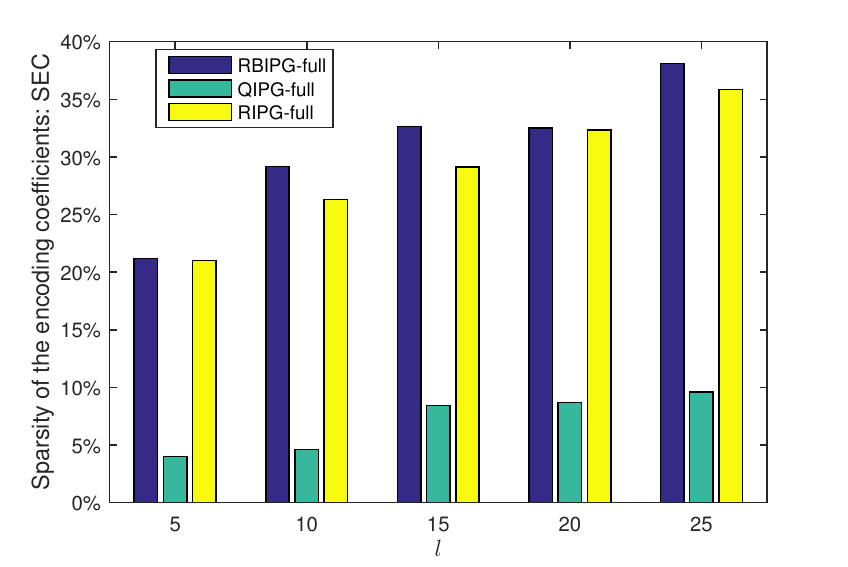}
	}
	\subfigure[The `full' case for Scenario (d)]{
		\includegraphics[width=7cm,height=4cm]{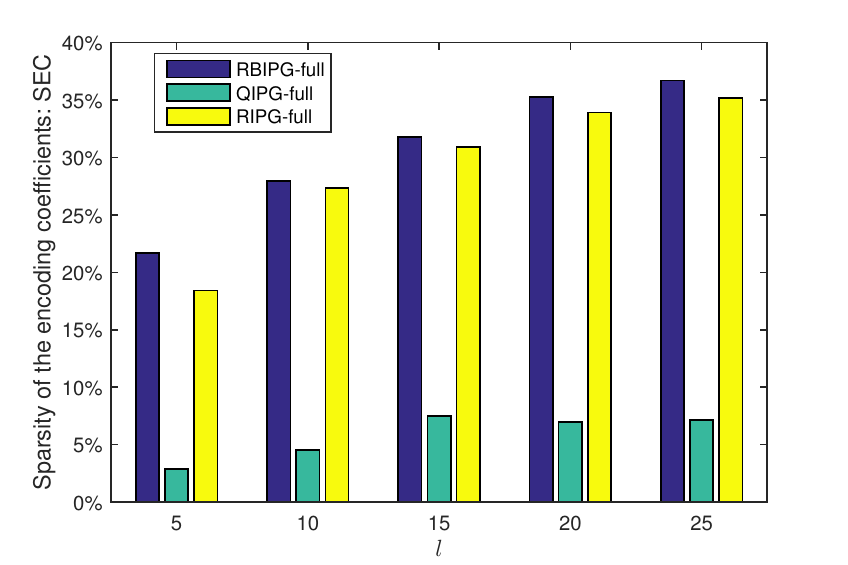}
	}
	\caption{Comparison of SEC with different values of $l$ under experiments \ref{ar_exper} and \ref{KDEF_exper}.}
	\label{sparsityh1}
\end{figure*}

\begin{table}[htbp]
	\caption{The recognition accuracy of different methods under experiment \ref{ar_exper}.}
	\label{ar_qu}
	\centering
	\resizebox{8cm}{4cm}{
		\begin{tabular}{|c|ccccc|}		
			\hline
			$l$&$l=10$&$l=20$&$l=30$&$l=40$&$l=50$\\ 
			\hline	
			Methods: &\multicolumn{5}{c|}{Scenario (a)}\\
			\hline	
			RBIPG-full  &\textbf{54.43\%}&\textbf{72.00\%}&\textbf{74.29\%}&\textbf{79.20\%}&\textbf{83.57\%}\\
			\hline
			RBIPG-pure
			&51.43\%&66.71\%&72.86\%&77.43\%&81.57\%\\
			\hline
			QIPG-full &49.57\%&61.57\%&73.14\%&74.43\%&78.57\%\\
			\hline
			QIPG-pure &50.29\%&64.29\%&73.43\%&78.14\%&81.00\%\\
			\hline
			RIPG-full &52.71\%&66.00\%&73.00\%&76.00\%&79.57\%\\
			\hline
			RIPG-pure &53.43\%&66.29\%&73.14\%&75.71\%&78.86\%\\
			\hline 
			RIPG-full$^{v}$&46.00\%&62.43\%&70.71\%&74.14\%&76.43\%\\
			\hline
			RIPG-pure$^{v}$&44.86\%&62.43\%&69.14\%&75.14\%&77.57\%\\
			\hline
			QPCA-full &51.57\%&59.57\%&64.14\%&66.71\%&67.71\%\\
			\hline
			QPCA-pure &52.43\%&60.71\%&64.71\%&67.29\%&68.29\%\\
			\hline
			Methods: &\multicolumn{5}{c|}{Scenario (b)}\\
			\hline	
			RBIPG-full  &\textbf{42.92\%}&\textbf{55.08\%}&\textbf{59.85\%}&\textbf{64.31\%}&\textbf{66.23\%}\\
			\hline
			RBIPG-pure
			&40.54\%&53.38\%&56.85\%&60.12\%&63.46\%\\
			\hline
			QIPG-full &35.23\%&48.92\%&55.46\%&61.38\%&63.38\%\\
			\hline
			QIPG-pure &39.38\%&51.69\%&56.31\%&62.15\%&64.54\%\\
			\hline
			RIPG-full &42.23\%&52.15\%&57.23\%&62.23\%&64.46\%\\
			\hline
			RIPG-pure &41.92\%&52.69\%&57.31\%&62.85\%&64.38\%\\
			\hline
			RIPG-full$^{v}$ &33.54\%&50.08\%&52.38\%&58.31\%&61.15\%\\
			\hline
			RIPG-pure$^{v}$ &35.54\%&46.31\%&55.15\%&59.69\%&61.62\%\\
			\hline
			QPCA-full &40.92\%&48.85\%&52.85\%&55.00\%&56.38\%\\
			\hline
			QPCA-pure &41.69\%&50.00\%&53.38\%&55.54\%&57.08\%\\
			\hline
	\end{tabular}}
\end{table}

\subsubsection{Color Face Recognition for KDEF Database \cite{lundqvist1998karolinska}}\label{KDEF_exper}
This experiment employs the KDEF database, containing 4,900 facial images from 70 subjects (35 female/35 male) displaying seven emotions (Angry, Fearful, Disgusted, Sad, Happy, Surprised, Neutral), each captured from five angles in two recording sessions (A/B series). Figure \ref{fig_face_ar}(b) displays samples of one subject.
We set up the experiments in the following two scenarios:
\begin{enumerate}
	\item [(c)] The training set comprises $17$ randomly selected faces from series A, with the remaining $18$ faces forming the test set, all resized to $80 \times 100$ pixels.
	\item [(d)] The training set contains $12$ randomly chosen faces from series B, with $23$ faces reserved for testing, all resized to $80 \times 100$ pixels.
\end{enumerate}

\begin{table}[htbp]
	\caption{The recognition accuracy of different methods under experiment \ref{KDEF_exper}.}
	\label{KDEF_qu}
	\centering
	\resizebox{8cm}{4cm}{
		\begin{tabular}{|c|ccccc|}		
			\hline
			$l$&$l=5$&$l=10$&$l=15$&$l=20$&$l=25$\\ 
			\hline	
			Methods: &\multicolumn{5}{c|}{Scenario (c)}\\
			\hline	
			RBIPG-full  &\textbf{77.38\%}&\textbf{90.71\%}&\textbf{92.38\%}&\textbf{93.81\%}&\textbf{93.89\%}\\
			\hline
			RBIPG-pure
			&76.59\%&85.08\%&90.71\%&91.75\%&92.46\%\\
			\hline
			QIPG-full &67.06\%&87.46\%&89.13\%&91.03\%&92.86\%\\
			\hline
			QIPG-pure &67.94\%&87.86\%&90.79\%&92.14\%&93.10\%\\
			\hline
			RIPG-full &71.35\%&88.33\%&90.79\%&91.90\%&92.94\%\\
			\hline
			RIPG-pure &72.54\%&88.17\%&90.79\%&91.59\%&92.70\%\\
			\hline
		    RIPG-full$^{v}$&56.75\%&84.84\%&87.78\%&90.71\%&92.30\%\\
			\hline
			RIPG-pure$^{v}$&61.43\%&85.71\%&90.63\%&91.11\%&91.59\%\\
			\hline
			QPCA-full &68.65\%&84.44\%&89.84\%&91.19\%&92.54\%\\
			\hline
			QPCA-pure &69.92\%&85.40\%&90.40\%&91.59\%&92.78\%\\
			\hline
			Methods: &\multicolumn{5}{c|}{Scenario (d)}\\
			\hline	
			RBIPG-full  &\textbf{69.81\%}&\textbf{82.86\%}&\textbf{84.29\%}&\textbf{86.40\%}&\textbf{88.70\%}\\
			\hline
			RBIPG-pure
			&63.73\%&79.13\%&82.11\%&85.34\%&87.20\%\\
			\hline
			QIPG-full &57.20\%&76.96\%&80.81\%&83.66\%&86.46\%\\
			\hline
			QIPG-pure &57.02\%&78.51\%&81.86\%&84.41\%&87.27\%\\
			\hline
			RIPG-full &61.86\%&78.57\%&82.67\%&85.34\%&86.89\%\\
			\hline
			RIPG-pure &63.85\%&78.57\%&82.24\%&85.59\%&87.33\%\\
			\hline
			RIPG-full$^{v}$&48.63\%&75.47\%&80.06\%&82.55\%&85.28\%\\
			\hline
			RIPG-pure$^{v}$&49.69\%&73.48\%&81.24\%&83.11\%&83.66\%\\
			\hline
			QPCA-full &60.56\%&76.40\%&82.24\%&84.60\%&85.65\%\\
			\hline
			QPCA-pure &62.05\%&77.27\%&82.55\%&85.16\%&86.27\%\\
			\hline
	\end{tabular}}
\end{table}

Figure \ref{rse} presents the reconstruction residual (RES) versus iterations for RBIPG, QIPG, and RIPG under experiment \ref{ar_exper}, Scenario (a), $l=50$ (for other values of $l$, similar results can be obtained). For RBIPG,
${\rm{RES}}=\|\ddot{\mathbf{X}}-\ddot{\mathbf{W}}\ddot{\mathbf{H}}\|_{F}$; for QIPG, ${\rm{RES}}=\|\dot{\mathbf{X}}-\dot{\mathbf{W}}\dot{\mathbf{H}}\|_{F}$; for RIPG-full, ${\rm{RES}}=(\sum_{s=0}^{3}\|\mathbf{X}_{s}-\mathbf{W}_{s}\mathbf{H}_{s}\|_{F}^{2})^{1/2}$; for RIPG-pure, ${\rm{RES}}=(\sum_{s=1}^{3}\|\mathbf{X}_{s}-\mathbf{W}_{s}\mathbf{H}_{s}\|_{F}^{2})^{1/2}$; for RIPG-full$^{v}$, ${\rm{RES}}=\|\mathbf{X}^{full}-\mathbf{W}\mathbf{H}\|_{F}$, and for RIPG-pure$^{v}$, ${\rm{RES}}=\|\mathbf{X}^{pure}-\mathbf{W}\mathbf{H}\|_{F}$\footnote{The columns of 
$\mathbf{X}^{full}$ represent flattened four-channel face images (including $\mathbf{X}_{av}$), while $\mathbf{X}^{pure}$ contains flattened three-channel color images.}. From Figure \ref{rse}, we can observe that these algorithms all tend to converge within $400$ iterations. After convergence, the sizes of the RES are relatively close, with the RES of RBIPG being slightly higher than those of QIPG and RIPG, except for RIPG-full$^{v}$ and RIPG-pure$^{v}$.
However, due to the deficiencies of the QNQMF model and the fact that the NMF model does not consider the potential relationships between color channels, the slightly lower RES value does not imply that they can achieve better basis and encoding coefficient matrices; they may be overfitting. The RES for RIPG-full$^{v}$ and RIPG-pure$^{v}$ are the highest, which may be due to the aggressive vectorization of all color channels, severely disrupting the data structure and the underlying low-rank prior.

Figure \ref{bi_RB_Q_ADMM1} and Figure \ref{bi_RB_Q_ADMM2} respectively show the basis matrices of RBIPG, QIPG, and RIPG\footnote{The cases for RIPG-full$^{v}$ and RIPG-pure$^{v}$ are omitted from Figures \ref{bi_RB_Q_ADMM1} and \ref{bi_RB_Q_ADMM2}, as the interpretability of the $\mathbf{W}$ and $\mathbf{H}$ obtained under these conditions is lower.} under the `full' and `pure' settings in Scenario (a) of experiment \ref{ar_exper} with $l = 50$. Figure \ref{bi_RB_Q_ADMM3} shows the encoding coefficients of RBIPG, QIPG, and RIPG under the `full' setting in Scenario (a) of experiment \ref{ar_exper} with $l = 50$. To measure the sparsity of the encoding coefficients (SEC),  we define SEC as follows:
\begin{equation}\label{sec}\small
	\text{SEC} = \frac{\text{num}_1(\mathbf{X})}{\text{num}(\mathbf{X})} \times 100\%,
\end{equation} 
where ${\rm{num}}(\mathbf{X})$ returns the total number of elements in $\mathbf{X}$, while ${\rm{num_1}}(\mathbf{X})$ returns the number of elements in $\mathbf{X}$ with a modulus less than $10^{-5}$. For fair comparison, Figure \ref{sparsityh1} shows the channel-wise average SEC obtained by RBIPG, QIPG, and RIPG under the
`full' setting. Specifically, for RBIPG, the SEC is calculated as $\big(\frac{\text{num}_1({\rm{Re}}(\ddot{\mathbf{H}}))}{\text{num}({\rm{Re}}(\ddot{\mathbf{H}}))} \times 100\%+\frac{\text{num}_1({\rm{Im}}_{j}(\ddot{\mathbf{H}}))}{\text{num}({\rm{Im}}_{j}(\ddot{\mathbf{H}}))} \times 100\%\big)/2$, and QIPG and RIPG follow analogous computation procedures. Tables \ref{ar_qu} and \ref{KDEF_qu} present the face recognition accuracy of different methods in experiments \ref{ar_exper} and \ref{KDEF_exper}. From these experimental results, we mainly observe the following points:
\begin{enumerate}
	\item [1)] Comparing the results of RBIPG-full and RBIPG-pure, we notice that in Figure \ref{bi_RB_Q_ADMM2}(a), a significant number of features in components ${\rm{Re}}(\ddot{\mathbf{W}})$ and ${\rm{Im}}_{j}(\ddot{\mathbf{W}})$ have disappeared compared to Figure \ref{bi_RB_Q_ADMM1}(a). This observation is consistent with the analysis in Remark \ref{remark1}, as under non-negative constraints, the near-orthogonality between the basis and the encoding coefficients often arises from approximately disjoint supports (\emph{i.e.}, non-overlapping non-zero patterns), which indicates that the full setting (\ref{face_re}) is indeed more suitable for our model. The comparison of face recognition accuracy between them in Tables \ref{ar_qu} and \ref{KDEF_qu} also supports this conclusion. However, for QIPG, RIPG, and QPCA, the face recognition accuracy obtained with the `full' setting is not higher than that obtained with the `pure' setting. This is quite natural, as the introduction of $\mathbf{X}_{av}$ does not add additional feature information.
	\item [2)] Consistent with NMF principles, the derived basis images should demonstrate inherent sparsity. Figures \ref{bi_RB_Q_ADMM1} and \ref{bi_RB_Q_ADMM2} show that, compared to RBIPG and RIPG, the sparsity of the basis images obtained by QIPG is highly unbalanced. Specifically, the ${\rm{Re}}(\dot{\mathbf{W}})$ component of $\dot{\mathbf{W}}$ obtained by QIPG is not sparse at all (we use ${\rm{abs}}({\rm{Re}}(\dot{\mathbf{W}}))$ because ${\rm{Re}}(\dot{\mathbf{W}})$ contains negative numbers), even though components ${\rm{Im}}_{\mathbf{i}}(\dot{\mathbf{W}})$, ${\rm{Im}}_{\mathbf{j}}(\dot{\mathbf{W}})$, and ${\rm{Im}}_{\mathbf{k}}(\dot{\mathbf{W}})$ appear to be very sparse. Such inexplicable result may be due to the inherent shortcomings of the QNQMF model. In addition, it can be observed that the basis images obtained by our RBIPG method are slightly sparser compared to those obtained by the RIPG method. For Figures \ref{bi_RB_Q_ADMM1} and \ref{bi_RB_Q_ADMM2}, by simply calculating the proportion of zero elements (\emph{i.e.}, sparsity) in all the basis images obtained by each method, we get: RBIPG-full\footnote{Although the basis images obtained by RBIPG-pure are noticeably sparser, based on point $1)$ and the analysis in Remark \ref{remark1}, we primarily focus on the results of RBIPG-full.} at $33.86\%$, QIPG-full at $28.89\%$, RIPG-full at $29.55\%$, QIPG-pure at $20.85\%$, and RIPG-pure at $29.42\%$. Thus, our RBIPG method can obtain relatively sparser basis images\footnote{For other values of $l$ and data, we can obtain the same conclusion.}.
	\item [3)] The encoding coefficients represent the weights of basis images, and distinct facial features should lead to a certain level of sparsity. If such sparsity is absent, the decomposition model may be inadequate. From Figure \ref{bi_RB_Q_ADMM3}, it can be observed that the encoding coefficients obtained by RBIPG are sparser than those obtained by QIPG and RIPG. This conclusion is also clearly supported by the quantitative comparison in Figure \ref{sparsityh1}. Furthermore, from Figure \ref{bi_RB_Q_ADMM3}(b), it can be observed that the ${\rm{Re}}(\dot{\mathbf{H}})$ component of the QIPG encoding coefficients contains negative values, which is unreasonable and mainly stems from design flaws in the QNQMF model.
	\item [4)] From Tables \ref{ar_qu} and \ref{KDEF_qu}, it is clear that our RBIPG-full method has an advantage in color face recognition, which is consistent with our previous analysis and expectations. 
\end{enumerate}

\section{Conclusions}
\label{conclusion}
In the paper, we introduced the concept of a non-negative reduced biquaternion (RB) matrix and proposed a non-negative RB matrix factorization (NRBMF) model, taking advantage of the multiplication properties of RB matrices. The proposed NRBMF model effectively addressed the limitations of the recently introduced quasi non-negative quaternion matrix factorization (QNQMF) model, thereby providing a novel theoretical tool for color image processing. To solve the associated optimization problem, we developed an efficient RB projected gradient algorithm and provided a theoretical convergence analysis. Furthermore, the effectiveness and superiority of the proposed model were verified through numerical experiments on color face recognition. These initial results are promising and  provide a foundation for future work on both the theoretical and methodological aspects of NRBMF. 

In the future, we will further consider improving the NRBMF model, such as developing a sparse NRBMF and studying more suitable error metrics for NRBMF (in addition to the least squares criterion) to better apply it to color face recognition. Additionally, we plan to apply this model or its improved versions to other color image processing tasks, such as color image denoising and color image restoration.

\bibliographystyle{IEEEtran}
\bibliography{references}

\end{document}